\documentclass[letterpaper]{article} % DO NOT CHANGE THIS
\usepackage{aaai25}  % DO NOT CHANGE THIS
\usepackage{times}  % DO NOT CHANGE THIS
\usepackage{helvet}  % DO NOT CHANGE THIS
\usepackage{courier}  % DO NOT CHANGE THIS
\usepackage[hyphens]{url}  % DO NOT CHANGE THIS
\usepackage{graphicx} % DO NOT CHANGE THIS
\usepackage{natbib}  % DO NOT CHANGE THIS AND DO NOT ADD ANY OPTIONS TO IT
\usepackage{caption} % DO NOT CHANGE THIS AND DO NOT ADD ANY OPTIONS TO IT
\usepackage{algorithm}
\usepackage{algorithmic}

\usepackage{newfloat}
\usepackage{listings}
\DeclareCaptionStyle{ruled}{labelfont=normalfont,labelsep=colon,strut=off} % DO NOT CHANGE THIS
\floatstyle{ruled}
\newfloat{listing}{tb}{lst}{}
\floatname{listing}{Listing}
\title{Bayesian Optimization for Unknown Cost-Varying Variable Subsets \\ with No-Regret Costs}
\author{
    Vu Viet Hoang \textsuperscript{\rm 1} \equalcontrib,
    Quoc Anh Hoang Nguyen \textsuperscript{\rm 1} \equalcontrib,
    Hung The Tran \textsuperscript{\rm 2} 
    \equalcontrib \thanks{Corresponding author.}\\
}
\affiliations{
    \textsuperscript{\rm 1}FPT Software AI Center\\
    HoangVV18@fpt.com, AnhNQH1@fpt.com \\
\textsuperscript{\rm 2}Hanoi University of Science and Technology \\
Hungtt@soict.hust.edu.vn 
}

\usepackage{amsfonts}
\usepackage{amsmath}
\usepackage{amsthm}
\theoremstyle{plain}
\newtheorem{theorem}{Theorem}

\newtheorem{lemma}[theorem]{Lemma}

\theoremstyle{definition}

\theoremstyle{remark}

\begin{document}

\maketitle

\begin{abstract}

Bayesian Optimization (BO) is a widely-used method for optimizing expensive-to-evaluate black-box functions. Traditional BO assumes that the learner has full control over all query variables without additional constraints. However, in many real-world scenarios, controlling certain query variables may incur costs. Therefore, the learner needs to balance the selection of informative subsets for targeted learning against leaving some variables to be randomly sampled to minimize costs. This problem is known as \textit{Bayesian Optimization with cost-varying variable subsets} (BOCVS). While the goal of BOCVS is to identify the optimal solution with minimal cost, previous works have only guaranteed finding the optimal solution without considering the total costs incurred. Moreover, these works assume precise knowledge of the cost for each subset, which is often unrealistic.  In this paper, we propose a novel algorithm for the extension of the BOCVS problem with random and unknown costs that separates the process into exploration and exploitation phases. The exploration phase will filter out low-quality variable subsets, while the exploitation phase will leverage high-quality ones. Furthermore, we theoretically demonstrate that our algorithm achieves a sub-linear rate in both quality regret and cost regret, addressing the objective of the BOCVS problem more effectively than previous analyses. Finally, we show that our proposed algorithm outperforms comparable baselines across a wide range of benchmarks.

\end{abstract}

\section{Introduction}

Bayesian optimization (BO) is a powerful technique designed for the sample-efficient optimization of expensive-to-evaluate black-box objective functions. It has been effectively used in various important experimental design problems, such as hyperparameter optimization \cite{Snoek2012}, chemical synthesis \cite{ShieldsSLPDAJAD21}, and particle accelerator control \cite{Kirschner2022}. Typically, BO has access to all query variables of the objective function, allowing it to re-evaluate any variable without additional hamper. However, in many practical situations, not all query variables can be easily modified. Some variables may follow specific distributions, and altering their values deterministically may incur costs. This setting is known as \textit{Bayesian Optimization with Cost-varying variable subsets} (BOCVS), as introduced in \cite{NEURIPS2023_090b23d5}. The aim is to identify optimal candidates for the objective function while minimizing incurred costs as much as possible. Naturally, greater control over the query variables leads to higher costs. Therefore, in each iteration, the learner must balance between the set of controllable query variables and the associated costs. 

We give an illustrative example with the BOCVS setting. Other concrete scenarios can be found in \cite{NEURIPS2023_090b23d5}. In the context of adaptive control for manufacturing systems \cite{KRUGER2011783}, the goal is to optimize the production process to maximize output quality while managing the associated costs of process adjustments. The operator must adjust various control parameters, such as machine settings, temperature, and operational schedules, to enhance production quality. However, these adjustments can incur costs like energy consumption, maintenance or operating costs. To mitigate these costs, the operator can selectively modify certain control parameters while allowing others to vary randomly within a controlled range. The primary challenge lies in the fact that the cost constraints are not fully known or predictable, complicating the evaluation of the impact of randomized versus deterministic adjustments. These costs are often uncertain because they depend on fluctuating factors such as labor, or machine conditions. The operator must strike a balance between managing the risk of high, unknown costs and maintaining acceptable production quality.

To the best of our knowledge, UCB-CVS \cite{NEURIPS2023_090b23d5} is the first and only work that addresses this problem precisely. While UCB-CVS has shown promising theoretical and empirical results, there are some drawbacks. Firstly, the definition of their objective function focuses solely on finding the optimal solution and neglects the cost constraints. To incorporate cost constraints, the authors introduced the cost-varying cumulative regret, a cost-weighted version of the regular cumulative regret. While it can be proven that UCB-CVS achieves sublinear cost-varying cumulative regret, it is unclear whether it minimizes the total cost. For example, if there exists two control sets with similar performance while the cost is significantly difference, the sublinear rate for cost-varying regret stated in Theorem 4.1 \cite{NEURIPS2023_090b23d5} does not emphasize the advantage of choosing cheaper control set. Therefore, we need a new definition of the objective function and regret to better reflect the goals of the BOCVS setting. Secondly, UCB-CVS assumes that the cost of every control set is deterministically known, which is a strict assumption that limits the algorithm's practicality. It would be more desirable to have an algorithm that does not require precise knowledge of the cost of every control sets. Thirdly, UCB-CVS uses an $\epsilon$-schedule as a relaxation to enable exploration with cheaper control sets. However, choosing an appropriate $\epsilon$-schedule is challenging. In practice, UCB-CVS uses a simple \textit{explore-then-commit} approach instead of an $\epsilon$-schedule. This creates a mismatch between the theoretical algorithm and its practical implementation.

In this paper, we address an extension of the BOCVS problem where the cost of playing a subset is unknown. We introduce an efficient novel approach to analyzing the effectiveness of an algorithm by defining both quality regret and cost regret, which is a more rigorous analysis compared to previous work. An efficient algorithm should aim to minimize both types of regret. Consequently, we develop an algorithm designed to minimize these regrets simultaneously. The main contributions of this work are as follows:

\begin{itemize}
    \item We introduce an extension of the BOCVS problem in which the costs of the subset may be random and unknown.
    \item We develop a cost-sensitive algorithm for the extension version. We derive an upper bound on the cumulative regret for our algorithm and theoretically show that the algorithm achieves sub-linear cumulative regret of both the objective function value and the cost.
    \item We conducted an empirical evaluation of our proposed algorithm's performance against baseline methods across a variety of experimental conditions. This included testing on both synthetic and real-world datasets, specifically a plant growth dataset and an airfoil self-noise dataset, which are relevant to the precision agriculture and advanced manufacturing applications discussed earlier.
\end{itemize}

\section{Preliminaries}
\subsection{Gaussian Process and Bayesian Optimization}

We consider a query set $\mathcal{X}$ and an objective function $f: \mathcal{X} \rightarrow \mathbb{R}$ where \(\mathcal{X} \subset \mathbb{R}^d\). The learner's goal is to find the query that maximizes $f$, denoted as $\mathbf{x}^*=\operatorname{argmax}_{\mathbf{x} \in \mathcal{X}} f(\mathbf{x})$. However, $f$ is a black-box function, meaning it is not available in a closed form and can only be evaluated by submitting a query $\mathbf{x}_t \in \mathcal{X}$ at each iteration $t$, and receiving a noisy observation $y_t=f\left(\mathbf{x}_t\right)+\xi_t$, where each $\xi_t$ is i.i.d. $\sigma$-sub-Gaussian noise with zero mean.

Gaussian processes \cite{williams2006gaussian} (GPs) are a popular choice as a surrogate model for optimizing $f$ in BO because they enable exact posterior inference: The GP posterior estimate of $f$ at any query $\mathbf{x} \in \mathcal{X}$ after $t$ iterations is a Gaussian with posterior mean and variance given by
\begin{equation}
    \begin{aligned}
    \mu_t(\mathbf{x}) &:= \mathbf{k}_t(\mathbf{x})^{\top}\left(\mathbf{K}_t + \lambda \mathbf{I}\right)^{-1} \mathbf{y}_t, \\
    \sigma_t^2(\mathbf{x}) &:= k(\mathbf{x}, \mathbf{x}) - \mathbf{k}_t(\mathbf{x})^{\top}\left(\mathbf{K}_t + \lambda \mathbf{I}\right)^{-1} \mathbf{k}_t(\mathbf{x})
    \label{eq:GP}
\end{aligned}
\end{equation}
where $\mathbf{y}_t:=\left(y_j\right)_{j=1}^t \in \mathbb{R}^t, k$ is a positive semidefinite kernel (covariance function), $\mathbf{k}_t(\mathbf{x}):=$ $\left(k\left(\mathbf{x}, \mathbf{x}_j\right)\right)_{j=1}^t \in \mathbb{R}^t, \mathbf{K}_t:=\left(k\left(\mathbf{x}_j, \mathbf{x}_{j^{\prime}}\right)\right)_{j, j^{\prime}=1}^t \in \mathbb{R}^{t \times t}$, and $\lambda$ is an algorithm parameter; if the noise is a Gaussian with variance $\sigma^2$, then the true posterior is recovered with $\lambda=\sigma^2$. The kernel $k$ is an important modeling choice as the GP posterior mean will reside in the reproducing kernel Hilbert space (RKHS) associated with $k$. For simplicity, we assume w.l.o.g. that $k\left(\mathbf{x}, \mathbf{x}^{\prime}\right) \leq 1$ for all $\mathbf{x}, \mathbf{x}^{\prime} \in \mathcal{X}$. Kernel $k$ affects the maximum information gain (MIG) defined as
$$
\gamma_T(\mathcal{X}):=\max _{\left\{\mathbf{x}_t\right\}_{t=1}^T \subseteq \mathcal{X}} 0.5 \log \left|\mathbf{I}+\lambda^{-1} \mathbf{K}_T\right| .
$$
The MIG characterizes the statistical complexity of a problem and plays an integral role in the theoretical analysis. For the commonly used squared exponential kernel, $\gamma_T(\mathcal{X})=\mathcal{O}\left((\log T)^{d+1}\right)$, while for the Matérn kernel with $\nu>1, \gamma_T(\mathcal{X})=\mathcal{O}\left(T^{d(d+1) /(2 v+d(d+1))}(\log T)\right)$ \cite{srinivas2009gaussian}. Importantly, $\gamma_T(\mathcal{X})$ is increasing in the volume of $\mathcal{X}$ \cite{srinivas2009gaussian}.

% An acquisition function is designed to suggest the next function evaluation point. Some examples of acquisition functions include Expected Improvement \cite{movckus1975bayesian}, Upper Confidence Bound \cite{srinivas2009gaussian,chowdhury2017kernelized}, Thompson sampling \cite{chowdhury2017kernelized}.

\subsection{Multi-armed bandits}

In more formal terms, the multi-armed bandit (MAB) problem can be described as follows: given a set of \( k \) arms (or actions), each arm \( i \) provides a reward from an unknown probability distribution. The goal is to develop a strategy or policy that maximizes the cumulative reward over a sequence of \( n \) trials. 

One effective approach to solving the MAB problem is the Upper Confidence Bound (UCB) algorithm. The UCB algorithm addresses the exploration-exploitation dilemma by using a principled method to select which arm to pull. The core idea is to choose the arm that maximizes the upper confidence bound of its estimated reward, thereby balancing the potential for high rewards and the uncertainty associated with each arm \cite{slivkins2019introduction}.

The UCB \cite{slivkins2019introduction} algorithm can be summarized in the following steps:
\begin{enumerate}
    \item \textbf{Initialization}: Start by playing each arm once to gather initial data.
    \item \textbf{Estimation}: After each round \( t \), calculate the estimated average reward for each arm \( i \), denoted as \( \hat{r}_i(t) \)
    \item  \textbf{Confidence Bound Calculation:} Compute the upper confidence bound for each arm \( i \), given by:
   \[
   r^{UCB}_i(t) = \hat{r}_i(t) + \sqrt{\frac{2 \ln T}{N_i(t)}}
   \]
   where \( N_i(t) \) is the number of times arm \( i \) has been played up to round \( t \), and \( \ln T \) is the natural logarithm of the current round number, which accounts for the total number of trials and ensures logarithmic growth.
\item \textbf{Selection:} Select the arm \( i \) with the highest \( r^{UCB}_i(t) \) value for the next trial.
\end{enumerate}

The term \( \sqrt{\frac{2 \ln T}{N_i(t)}} \) represents the confidence interval around the estimated reward \( \hat{\mu}_i(t) \). As more data is gathered for a particular arm (i.e., as \( N_i(t) \) increases), the confidence interval narrows, reflecting increased certainty about the arm's true reward distribution. Conversely, for arms that have been played fewer times, the wider confidence interval encourages exploration by allowing the possibility of higher potential rewards.
\section{Problem Definition}

We introduce the extension of BOCVS setting in this section. The problem involves a compact query set \(\mathcal{X} \subset \mathbb{R}^d\) and an objective function \(f: \mathcal{X} \rightarrow \mathbb{R}\) that resides in the Reproducing Kernel Hilbert Space (RKHS) associated with a kernel \(k\), with the RKHS norm constrained by an upper bound \(B\). Without loss of generality, we assume that \(\mathcal{X} = [0, 1]^d\). Let \([d] = \{1, 2, \ldots, d\}\) represent the indices of the \(d\) variables.

The learner has access to a collection \(\mathcal{I} \subseteq 2^{[d]}\) of control sets, indexed by \(1, 2, \ldots, m\), where \(m = |\mathcal{I}|\). Each control set \(i \in [m]\), denoted as \(\mathcal{I}_i \subseteq [d]\), specifies the subset of variables that the learner can choose to control. The complement of \(\mathcal{I}_i\), denoted as \(\overline{\mathcal{I}}_i = [d] \setminus \mathcal{I}_i\), indicates the subset of variables that will be randomly sampled from a known distribution.

A query \(\mathbf{x} \in \mathcal{X}\) can thus be decomposed into two parts: the control partial query \(\mathbf{x}^i\) and the random partial query \(\mathbf{x}^{-i}\). The control partial query is defined as \(\mathbf{x}^i = (x_\ell)_{\ell \in \mathcal{I}_i}\), which includes the variables indexed by \(\mathcal{I}_i\). The random partial query is defined as \(\mathbf{x}^{-i} = (x_\ell)_{\ell \in \overline{\mathcal{I}}_i}\), which includes the variables indexed by \(\overline{\mathcal{I}}_i\). Therefore, we can write the query $x \in \mathcal{X}$ as $x = [\mathbf{x}^i, \mathbf{x}^{-i}]$. Furthermore, let \(\mathcal{X}^i = \{\mathbf{x}^i \mid \mathbf{x} \in \mathcal{X}\}\) denote the set of all possible control partial queries.

%It is important to note that \([\mathbf{x}^i, \mathbf{x}^{-i}]\) does not represent a simple concatenation of vectors, as the variables may need to be reordered according to their indices to form the complete query \(\mathbf{x}\). Furthermore, let \(\mathcal{X}^i = \{\mathbf{x}^i \mid \mathbf{x} \in \mathcal{X}\}\) denote the set of all possible control partial queries.

In each iteration \( t \), the learner selects a control set \( i_t \in \mathcal{I} \) and specifies the values for the control partial query \(\mathbf{x}^{i_t}_t\). The values for the random partial query \(\mathbf{x}^{-i_t}_t\) are then sampled from the environment according to a known distribution. To simplify, the complete query for iteration \( t \) is \(\mathbf{x}_t = [\mathbf{x}_t^{i_t}, \mathbf{x}_t^{-i_t}] = [\mathbf{x}^{i_t}, \mathbf{x}^{-i_t}] = (x_{t, \ell})_{\ell \in [d]}\). For each variable \( x_{t, \ell} \) where \(\ell \in \overline{\mathcal{I}}_{i_t}\), it is a realization of a random variable \( X_{t, \ell} \sim \mathcal{P}_\ell \). Consequently, the observed random partial query \(\mathbf{x}^{-i_t}\) is a realization of the random vector \(\mathbf{X}^{-i_t} = (X_{t, \ell})_{\ell \in \overline{\mathcal{I}}_{i_t}} \sim \mathbb{P}^{-i_t}\), where \(\mathbb{P}^{-i_t}\) is the product measure \(\prod_{\ell \in \overline{\mathcal{I}}_{i_t}} \mathcal{P}_\ell\). This means each variable in the random partial query is independently sampled from its respective probability distribution. All distributions \(\mathcal{P}_\ell\) are assumed to be known. The learner then observes the output \( y_t = f(\mathbf{x}_t) + \xi_t \), where \(\xi_t\) is independently and identically distributed \(\sigma\)-sub-Gaussian noise with a zero mean.

The learner wishes to find the optimal control set $i^+$ and specified values in control partial query $\mathbf{x}^{i^+}$ that maximize the expected value of $f\left(\left[\mathbf{x}^i, \mathbf{X}^{-i}\right]\right)$ where the expectation is w.r.t. $\mathbf{X}^{-i} \sim \mathbb{P}^{-i}$ :
\begin{equation}
    \left(i^+, \mathbf{x}^{i^+}\right) \in \underset{\left(i, \mathbf{x}^i\right) \in[m] \times \mathcal{X}^i}{\operatorname{argmax}} \mathbb{E}\left[f\left(\left[\mathbf{x}^i, \mathbf{X}^{-i}\right]\right)\right] .
\end{equation}

In every iteration $t$, the learner pays a cost $c_{i_t}^{(t)}$, which sampled from a unknown distribution with mean $c_{i_t}$. The distribution of costs for each subset may be different. The learning procedure ends after $T$ iterations when $C-\sum_{t=1}^T c_{i_t}^{(t)}<c_{i_{T+1}}^{(t)}$, where $C$ is the budget that can be paid. Note that different from the original BOCVS setting \cite{NEURIPS2023_090b23d5}, we do not require to know exactly the cost of every control subsets or the ordering of cost for control subsets, i.e $c_1 \leq c_2 \leq \ldots \leq c_m.$

In some scenarios, it is acceptable for the learner to sacrifice the quality of the objective value in order to manage costs more efficiently. For instance, in the adaptive control problem for manufacturing systems \cite{KRUGER2011783}, it is more practical for the operator to keep production quality above a certain predefined threshold rather than constantly striving to maximize product quality, which would result in higher costs. Therefore, different from \cite{NEURIPS2023_090b23d5}, to manage effectively costs spent, assume $f(x) \geq 0$ (can be achieved by subtracting the worst value that $f$ can achieve), we allow the learner to be agnostic between subsets, whose expected reward is greater than $1 - \alpha$ fraction of the highest expected value, for a fixed and known value of $\alpha$. The learner’s objective is
to learn and exploit the cheapest subset among these high-quality subsets as frequently as possible to specify the optimal value for the corresponding partial control query. More specifically, we denote the set of subsets $\mathcal{C}_*$ whose expected value of $f\left(\left[\mathbf{x}^i, \mathbf{X}^{-i}\right]\right)$ is within $1-\alpha$ factor of the highest expected value, i.e, \begin{align*}
    \mathcal{C}_* = \Bigg\{ i \in \mathcal{I} \ \Bigg| \ & \exists \mathbf{x}_i \in \mathcal{X}_i :\mathbb{E}\left[f\left(\left[\mathbf{x}^i, \mathbf{X}^{-i}\right]\right)\right] \\
    &\geq (1-\alpha) \max_{\substack{(i', \mathbf{x}^{i'}) \\ \in [m] \times \mathcal{X}^{i'}}} \mathbb{E}\left[f\left(\left[\mathbf{x}^{i'}, \mathbf{X}^{-i'}\right]\right)\right] \Bigg\}
\end{align*}
The learner’s goal is to design an algorithm that will exploit the cheapest subset whose expected value is at least as large as the smallest tolerated reward. In other words, the learner needs to simultaneously identify and  maximize the number of choices of subsets $i_*=\operatorname{argmin}_{i \in \mathcal{C}_*} c_i .$
To measure the performance of any algorithms, we use two notions of regret: 
\begin{itemize}
 \item Quality regret:

$$
\begin{aligned}
    R^f_T:=\sum_{t=1}^T ((1-\alpha)&\mathbb{E}\left[f\left(\left[\mathbf{x}^{i^+}, \mathbf{X}^{-i^+}\right]\right)\right]\\ &-\mathbb{E}\left[f\left(\left[\mathbf{x}^{i_t}, \mathbf{X}^{-i_t}\right]\right)\right])
\end{aligned}
$$
\item Cost regret
$$
R^c_T:=\sum_{t=1}^T \max \{\left( c_{i_t} - c_{i^*}\right), 0\}
$$
\end{itemize}
The definitions of quality regret and cost regret in the paper show a notable resemblance to those outlined in \cite{SinhaSKA21}. In \cite{SinhaSKA21}, the authors employ the concepts of rewards and costs associated with arms in the MAB problem to characterize these types of regret. The objective then is to design an algorithm that simultaneously minimizes both the cost and quality regret.

\paragraph{Discussion}
% This paragraph has been revised.
Our definition of the objective function and regret introduces a new factor, $\alpha$, which represents the extent to which the learner is willing to forgo the quality of the objective value in order to reduce costs. When $\alpha$ approximates 0, the problem focuses solely on optimizing the objective function, and the cost can be neglected (if there are multiple optimal subsets, the focus is on finding the one with the smallest cost among them). In contrast, as $\alpha$ approaches 1, the cost is optimized more effectively, but the quality of the objective function solution can decrease with $T$ given. Therefore, the goal of optimizing both regrets is to find a balanced solution with the parameter $\alpha$. This makes our framework more generalized and applicable to a broader range of real-world scenarios. Furthermore, our regret analysis is more thorough and better aligned with the goals of BOCVS compared to previous work. \citet{NEURIPS2023_090b23d5} define the cost-varying cumulative regret for the BOCVS problem as follows:
$$
R_T:=\sum_{t=1}^T c_{i_t}\left(\mathbb{E}\left[f\left(\mathbf{x}^{i^*}, \mathbf{X}^{-i^*}\right)\right]-\mathbb{E}\left[f\left(\mathbf{x}^{i_t}, \mathbf{X}^{-i_t}\right)\right]\right)
$$
In the case where $\alpha = 0$, if an algorithm achieves sublinear $R_T^f$, it will also achieve sublinear $R_T$. However, the reverse is not true. When $\alpha = 0$ and multiple optimal subsets exist, an algorithm that achieves sublinear $R_T$ may not simultaneously achieve sublinear $R_T^f$ and $R_T^c$. Analyzing both types of regret provides a more comprehensive assessment of the algorithm's performance and its cost management.

\section{Method}
In this section, we introduce an explore-then-commit algorithm to address Bayesian Optimization with unknown cost-varying subsets. At the high-level idea, this algorithm initially explores each subset for a predetermined number of rounds before transitioning to the exploitation phase. During the exploitation phase, the algorithm identifies a feasible set of subsets based on the upper and lower confidence bounds (UCB, LCB) of their maximum values in each round. This set includes all subsets where the upper confidence bound of their maximum expected values exceeds the minimum acceptable value. From this feasible set, the subset with the lowest estimated cost is selected for exploitation. The detailed procedure is outlined in Algorithm \ref{alg}. A crucial part of this algorithm is determining the appropriate number of exploration rounds before moving to the exploitation phase, which will be discussed in the next section.

At iteration \( t \), the Gaussian Process (GP) posterior estimate of $f$ is used to define an upper confidence bound \( u_{t-1} \) and a lower confidence bound for \( l_{t-1} \) for \( f \), represented as:  
\[
u_{t-1}(\mathbf{x}) = \mu_{t-1}(\mathbf{x}) + \beta_t \sigma_{t-1}(\mathbf{x}),
\]  
\[
l_{t-1}(\mathbf{x}) = \mu_{t-1}(\mathbf{x}) - \beta_t \sigma_{t-1}(\mathbf{x}),
\]  
where \( \mu_{t-1}(\mathbf{x}) \) is the posterior mean, \( \sigma_{t-1}(\mathbf{x}) \) is the posterior standard deviation, and \( \beta_t \) is a parameter controlling the exploration-exploitation trade-off. The goal of the exploration phase is to filter out low-quality control subsets by setting an initial threshold for the exploitation phase. This is achieved by defining the intersected lower confidence bound $\overline{lcb}$ for $\mathbb{E}\left[f\left(\left[\mathbf{x}^{i^{+}}, \mathbf{X}^{-i^{+}}\right]\right)\right]$ and the intersected upper confidence bound $\overline{ucb}_i$ for $\max _{ \mathbf{x}^i \in \mathcal{X}^i} \mathbb{E}\left[f\left(\left[\mathbf{x}^i, \mathbf{X}^{-i}\right]\right)\right]$ in lines 10 and 11. High-quality control sets $i$ must meet the condition $\overline{ucb}_i \geq (1-\alpha)\overline{lcb}$, forming the set $\mathcal{S}_1$ that includes only the high-quality subsets. Additionally, during the exploitation phase, the intersected confidence bounds $\overline{lcb}$ and $\overline{ucb}_i$ are updated for each control subset $i$ in lines 14, 15, 18, and 19. This update mechanism ensures that, with high probability, the set $\mathcal{S}_1$ gradually narrows over time while always containing the optimal set $\mathcal{C}_*$. As a result, the condition $\mathcal{S}_1 = \emptyset$ in line 17 is likely to occur with low probability. However, if this condition does arise, the intersected confidence bounds is updated so that $\mathcal{S}_1$ is revised in line 20 to ensure it is not empty. Without considering cost regret, selecting any subset within $\mathcal{S}_1$ would provide satisfactory performance by minimizing quality regret. However, to optimize both cost and quality regret, it is essential to distinguish between the subsets and identify the one with the lowest cost in set $\mathcal{S}_1$. Since the exact cost is unknown, it is approximated using $c_i^{\text{LCB}}$, as defined in line 24. This approximation is then used to balance between exploring various subsets and exploiting the currently cheapest subset. Finally, the promising subset is selected, and the values in the control partial query are specified in lines 25 and 26.

\begin{algorithm}[h!]
\caption{Proposed method}
    \begin{algorithmic}[1]
        \STATE \textbf{Input:} $T$ evaluations, GP with kernel $k$, control sets $\mathcal{I}$, costs $\left(c_i\right)_{i=1}^m$, $\tau$, tolerated parameter $\alpha$
\STATE \textbf{Exploration phase:}
\FOR{for $t \in[1, m \tau]$}
\STATE $i:=t \bmod m ;$
\STATE $x^i_t := \arg \max _{\mathbf{x}^i \in \mathcal{X}^i} \mathbb{E}\left[u_{t-1}\left(\left[\mathbf{x}^i, \mathbf{X}^{-i}\right]\right)\right]$
\STATE Observe $\mathbf{x}^{-i_t}$ drawn from $\mathbb{P}^{-i_t}$
\STATE Observe $y_t:=f\left(\mathbf{x}_t^{i}\right)+\xi_t$ and $c_i^{(t)}$
\STATE $T_i(t+1):=T_i(t)+\mathbf{1}\left\{I_t=i\right\} \forall i \in[m]$
\ENDFOR
\STATE $\overline{lcb}:=\underset{\left(i, \mathbf{x}^i\right) \in[m] \times \mathcal{X}^i, t\leq m\tau}{\max} \mathbb{E}\left[l_{t-1}\left(\left[\mathbf{x}^i, \mathbf{X}^{-i}\right]\right)\right]$
\STATE $\overline{ucb}_i:=\underset{k\leq m}{\min} \underset{ \mathbf{x}^i \in\mathcal{X}^i}{\max} \mathbb{E}\left[u_{k\tau+i-1}\left(\left[\mathbf{x}^i, \mathbf{X}^{-i}\right]\right)\right], \forall i \in [m]$

\STATE \textbf{Exploitation phase:}
\FOR{ iteration $t=m\tau + 1$ to $T$}
\STATE $\overline{lcb} := \max \left\{ \overline{lcb}, \underset{\left(i, \mathbf{x}^i\right) \in[m] \times \mathcal{X}^i}{\max} \mathbb{E}\left[l_{t-1}\left(\left[\mathbf{x}^i, \mathbf{X}^{-i}\right]\right)\right] \right\} $
\STATE $\overline{ucb}_i := \min \left\{ \overline{ucb}_i, \underset{ \mathbf{x}^i \in\mathcal{X}^i}{\max} \mathbb{E}\left[u_{t-1}\left(\left[\mathbf{x}^i, \mathbf{X}^{-i}\right]\right)\right] \right\}$, $\forall i \in [m] $
\STATE $\mathcal{S}_1:=\left\{i \in [m] \mid \overline{ucb}_i  > (1 - \alpha) \overline{lcb} \right\}$
\IF{$\mathcal{S}_1 = \emptyset $}
\STATE $\overline{lcb} := \max _{\left(i, \mathbf{x}^i\right) \in[m] \times \mathcal{X}^i} \mathbb{E}\left[l_{t-1}\left(\left[\mathbf{x}^i, \mathbf{X}^{-i}\right]\right)\right]$
\STATE $\overline{ucb}_i := \max _{ \mathbf{x}^i \in\mathcal{X}^i} \mathbb{E}\left[u_{t-1}\left(\left[\mathbf{x}^i, \mathbf{X}^{-i}\right]\right)\right]$, $\forall i \in [m] $
\STATE $\mathcal{S}_1:=\left\{i \in [m] \mid \overline{ucb}_i  > (1 - \alpha) \overline{lcb} \right\}$
\ENDIF
\STATE $\hat{c}_i(t):=\frac{\sum_{k=1}^{t-1} c_i^{(k)}\left\{I_k=i\right\}}{T_i(t)} \forall i \in[m] ;$
\STATE $\beta_i(t) := \sqrt{\frac{2 \log T}{T_i(t)}} \forall i \in[m] ;$
\STATE $c_i^{\mathrm{LCB}}(t) := \max \left\{\hat{c}_i(t)-\beta_i(t), 0\right\} \forall i \in[m]$
\STATE $\mathcal{S}_3:=\left\{i \in \mathcal{S}_1 \mid c_i^{\mathrm{LCB}}=\min _{j \in \mathcal{S}_1} c_j^{\mathrm{LCB}}\right\}$
\STATE $\left(i_t, \mathbf{x}^{i_t}\right):=\underset{\left(i, \mathbf{x}^i\right) \in \mathcal{S}_3 \times \mathcal{X}^i}{\operatorname{argmax}} \mathbb{E}\left[u_{t-1}\left(\left[\mathbf{x}^i, \mathbf{X}^{-i}\right]\right)\right]$

\STATE Observe $\mathbf{x}^{-i_t}$ drawn from $\mathbb{P}^{-i_t}$
\STATE Observe $y_t:=f\left(\mathbf{x}_t^{i}\right)+\xi_t$ and $c_{i_t}^{(t)}$
\STATE Update $\mathcal{D}_t:=\left\{\left(\mathbf{x}_\tau, y_\tau\right)\right\}_{\tau=1}^t$
\ENDFOR
\STATE \textbf{return} $\mathcal{D}_t$
    \end{algorithmic}
    \label{alg}
\end{algorithm}

% \section{Problem}
% The learner wishes to find the optimal control set $i^*$ and specified values in control partial query $\mathbf{x}^{i^*}$ that maximize the expected value of $f\left(\left[\mathbf{x}^i, \mathbf{X}^{-i}\right]\right)$ where the expectation is w.r.t. $\mathbf{X}^{-i} \sim \mathbb{P}^{-i}$ :
% $$
% \left(i^*, \mathbf{x}^{i^*}\right):=\underset{\left(i, \mathbf{x}^i\right) \in[m] \times \mathcal{X}^i}{\operatorname{argmax}} \mathbb{E}\left[f\left(\left[\mathbf{x}^i, \mathbf{X}^{-i}\right]\right)\right] .
% $$

% The learner has an initial budget $C \in \mathbb{R}^{+}$and every control set $\mathcal{I}_i$ has an associated cost $c_i>0$ for all $i \in[m]$. Let the control set indices be defined such that $c_1 \leq c_2 \leq \ldots \leq c_m$.

% \newpage
% \paragraph{Regret Analysis.}
% We denote the optimal control set as \(i^*\) and the specified values in the control partial query as \(x_{i^*}\), which maximize the expected value. The cost of the control set \(i^*\) is denoted as \(c_{i^*}\). In the case of multiple optimal subsets, the set \(i^*\) is the optimal subset with the minimal cost.
% \begin{itemize}
%  \item Quality regret
% $$
% R^f_T:=\sum_{t=1}^T \left(\mathbb{E}\left[f\left(\left[\mathbf{x}^{i^*}, \mathbf{X}^{-i^*}\right]\right)\right]-\mathbb{E}\left[f\left(\left[\mathbf{x}^{i_t}, \mathbf{X}^{-i_t}\right]\right)\right]\right)
% $$
% \item Cost regret
% $$
% R^c_T:=\sum_{t=1}^T \max \{\left( c_{i_t} - c_{i^*}\right), 0\}
% $$
% \end{itemize}

\section{Main theorem}
\begin{theorem}
    Assume that the distribution of the random cost of each subset has support [0,1], $B$ is the upper bound of the RKHS norm of $f$, $M^{f}=\max _{x \in \mathcal{X}}\left\|f_x\right\|$. With probability at least $1-3\delta-2/T^2$, Alg. 1 incurs an objective cumulative regret and a cost cumulative regret bounded by 
    \begin{equation}
    \begin{aligned}
        R^f_T &= \mathcal{O} \left((m-1) \tau M^f \right)+ \\
    &\mathcal{O}\left(\beta_t \left(\sqrt{4(T+2-m \tau) \gamma_{T-m \tau}(\mathcal{X})}+m\left(\log \frac{m}{\delta}\right)\right)\right)\\
        &+ \mathcal{O}\left(\frac{T-(m-1) \tau}{\tau}\beta_{m\tau}\left(\sqrt{\tau \gamma_{\tau}\left(\mathcal{X}\right)}+\log \frac{m}{\delta}\right)\right)
    \end{aligned}  
    \end{equation}

\begin{equation}
\raggedleft
    R^c_T = \mathcal{O}(m\tau) + \mathcal{O}\left((T-m \tau) \sqrt{\frac{2 \log T}{\tau}}\right)
\end{equation}
by setting $\beta_t=B+\sigma \sqrt{2\left(\gamma_{t-1}(\mathcal{X})+1+\log (1 / \delta)\right)}$.
\label{theo:1}
\end{theorem}

For any appropriately chosen kernel such that $\gamma_T(\mathcal{X}) \leq \mathcal{O}(T^{\frac{1}{2} - \epsilon})$ where $\epsilon > 0$ (e.g., commonly used squared exponential kernel) and we choose $\tau$ satisfying that $\mathcal{O}(T^{\epsilon_0})\leq \tau < \mathcal{O}(T)$ where $\epsilon_0>0$, Theorem \ref{theo:1} implies that both the quality regret $R^f_T$ and the cost regret $R^c_T$ will be sublinear in $T$. The best control set and specified values in the control partial query in the algorithm’s choices eventually converge to the cheapest acceptable solution. This ensures the algorithm still achieves good performance while still managing costs effectively. The main idea of the theorem's proof is based on analyzing the components in different phases of the regrets. 

In case we explore subsets with a sufficiently large number of evaluations $\tau$, we can be assured that with high probability we will only focus on subsets in the set $\mathcal{C}_{*}$. We can describe this property by the following lemma:

\begin{lemma}
    If there exists a $\tilde{\epsilon}>0$ s.t. for all $i^\prime  \not \in \mathcal{C}_*, \mathbb{E}\left[f\left(\left[\mathbf{x}^{i^\prime}, \mathbf{X}^{-i^\prime}\right]\right)\right] \leq \left(1-\alpha\right)\max _{\mathbf{x}^i \in \mathcal{X}^i} \mathbb{E}\left[f\left(\left[\mathbf{x}^i, \mathbf{X}^{-i}\right]\right)\right]-\tilde{\epsilon}$, and $\gamma_T(\mathcal{X})<\mathcal{O}(\sqrt{T})$, with probability at least $1 - \delta$, if $\tau \geq \min \left\{t: \beta_{t m }\frac{\left(2 \sqrt{4(t+2) \gamma_{t}(\mathcal{X})} + 4 \log \frac{m}{\delta}+8 \log (4)+1\right)}{t} \leq \frac{\tilde{\epsilon}}{2-\sigma}\right\}$, then $\mathcal{S}_1 = \mathcal{C}_*$ for all iteration in Exploitation phase. 
    \label{lm:2}
\end{lemma}
This lemma’s proof is perhaps the trickiest and crucially relies on $\tau$ and distance gap $\tilde{\epsilon}$. From Lemma \ref{lm:2}, we can improve the upper bound of $R^f_T$ in theorem \ref{theo:1} as follows:
\begin{theorem}
    Suppose that the assumption of both Theorem \ref{theo:1} and Lemma \ref{lm:2} hold, with probability at least $1-3\delta-2/T^2$, Alg. 1 incurs an objective cumulative regret and a cost cumulative regret bounded by 
    \begin{equation}
\begin{aligned}
& R_T^f=\mathcal{O}\left(m \tau M^f\right)+ \\
& \mathcal{O}\left(\beta_t\left(\sqrt{4(T+2-m \tau) \gamma_{T-m \tau}(\mathcal{X})}+m\left(\log \frac{m}{\delta}\right)\right)\right) \\
& \quad R_T^c=\mathcal{O}(m \tau)+\mathcal{O}\left((T-m \tau) \sqrt{\frac{2 \log T}{\tau}}\right)
\end{aligned}
\end{equation}
\label{theo:2}
\end{theorem}
\paragraph{Practical Considerations} 
Theorems \ref{theo:1} and \ref{theo:2} give us an idea of how much evaluation is needed during the exploration phase depending on $T$ and $\tilde{\epsilon}$ for the algorithm to achieve the best performance. However, in practice, we are only given a budget without knowing these two parameters in advance. Therefore, we suggest using a larger budget for the exploration phase than for the exploitation phase. Moreover, if the application does not specify the tolerated parameter $\alpha$, we can allow it to decrease as the remaining budget nears zero. This gradual reduction allows us to work with many subsets that are sufficient to increase information gain while still ensuring low costs, before moving on to focus on playing better but more expensive subsets.
    
\section{Experiments}

\begin{figure*}
    \centering
    % \vspace{-0pt}
    \includegraphics[width=0.45\textwidth]{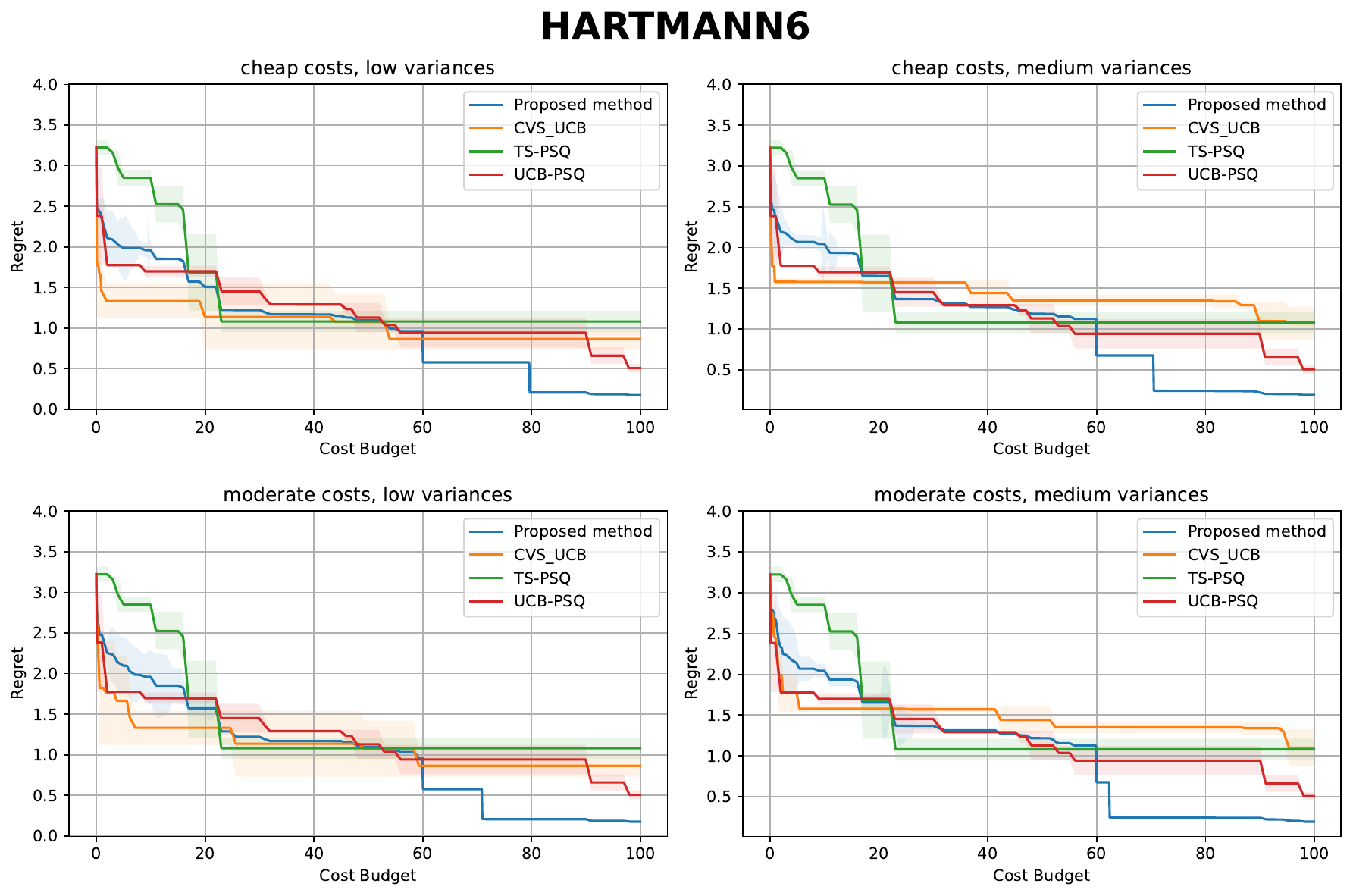}
    \includegraphics[width=0.45\textwidth]{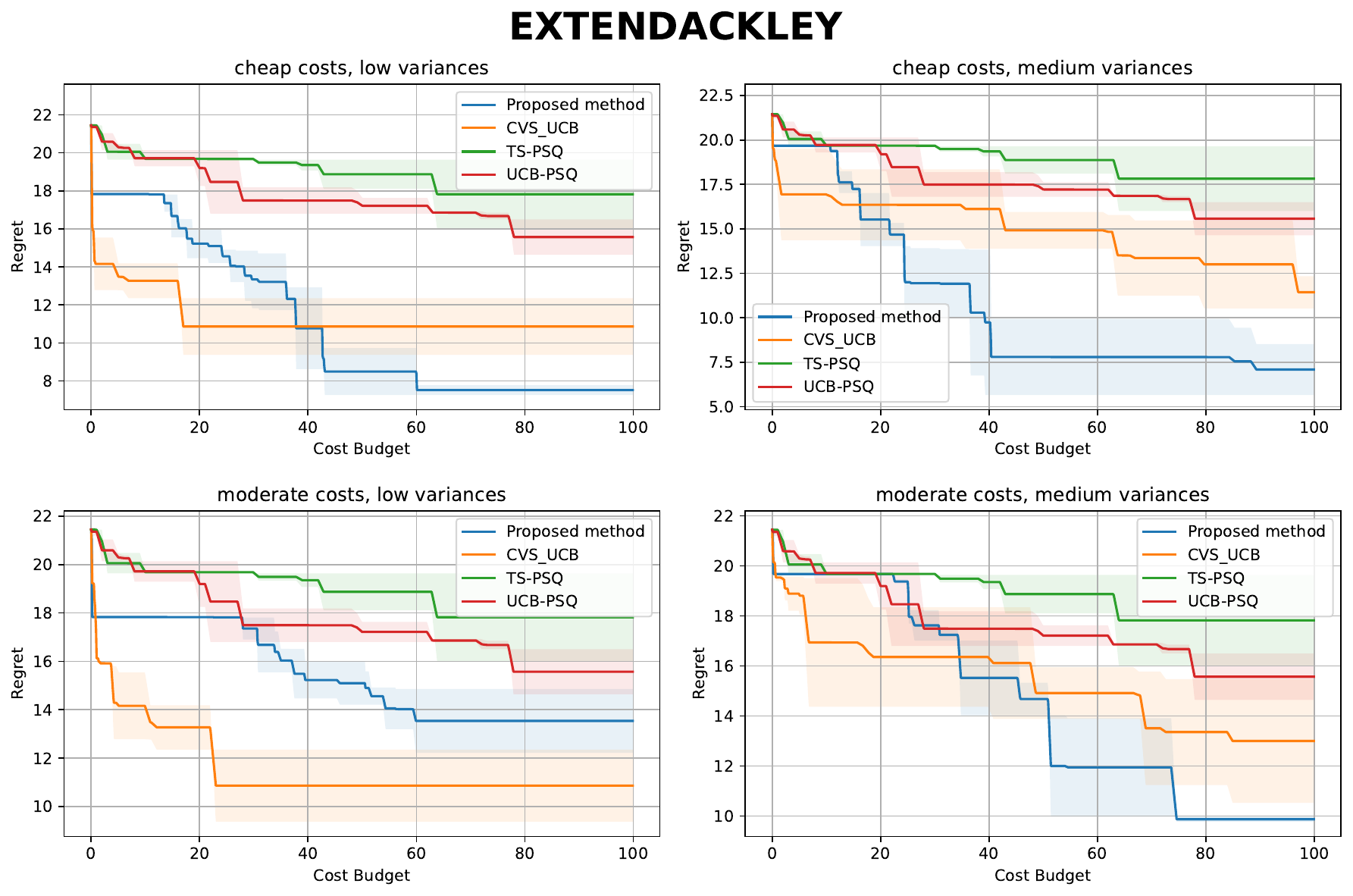}
    \includegraphics[width=0.45\textwidth]{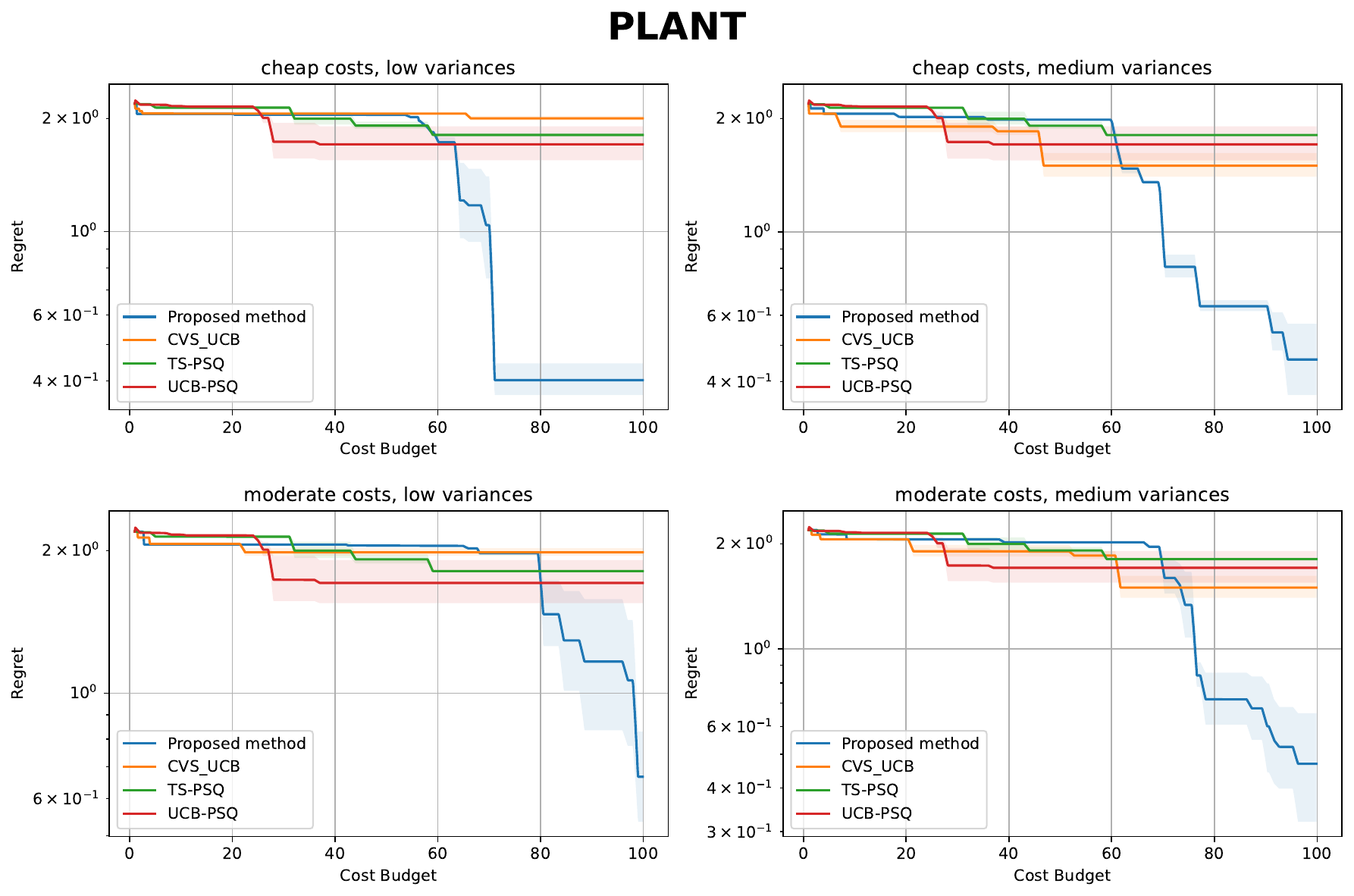}
    \includegraphics[width=0.45\textwidth]{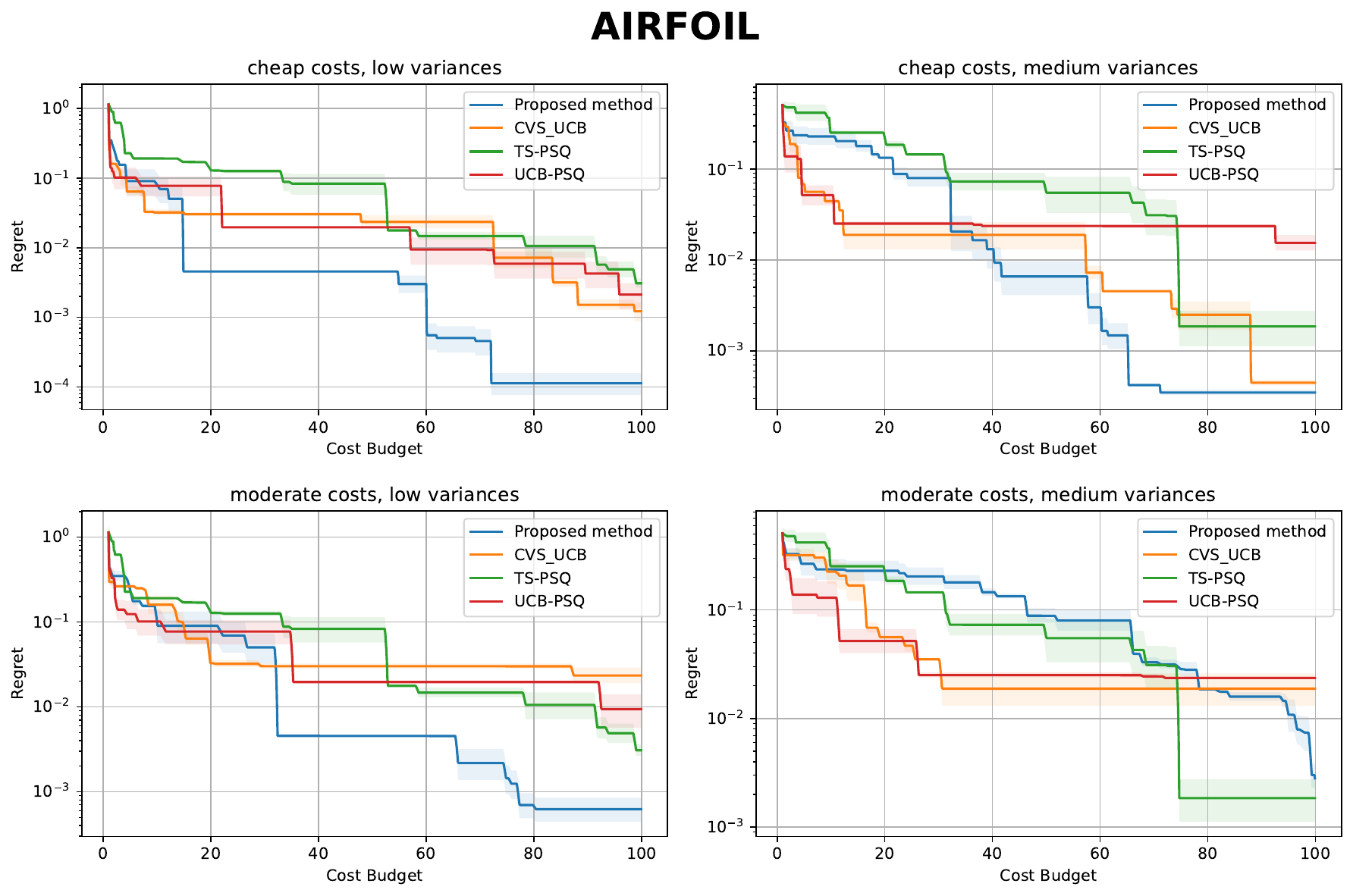}
    \caption{Comparison with the BO baselines.}
    \label{fig:main_results}
\end{figure*}

This section empirically evaluates the performance of the tested algorithms with 4 objective functions: (a) Hartmann $(d = 6)$ and (b) Ackley $(d = 6)$ as the synthetic benchmark functions, adding 6 unrelated variables, which leads to the existence of many subsets containing optimal points, (c) a plant growth simulator built from real-world data where the variables are nutrients such as $\mathrm{NH}_3$ and $\mathrm{pH}(5-$ D), and (d) a simulator built from the airfoil self-noise dataset (5-D) from the UCI Machine Learning Repository \cite{dua2017uci}. For the first two objective functions, there are 7 control sets, of which 2 are optimal sets: the full set and the subset containing valid variables. For the plant growth objective function, we pick 7 control sets including the full query control set. For the airfoil self-noise objective function, similar to that of \cite{hayashi2022bayesian}, we pick 7 control sets of 2 variables each that are not subsets of each other. We use 2 different sets of means of costs $c$ for the 7 control sets: cheap $(\{0.01,0.01,0.01,0.1,0.1,0.1,1\})$, moderate $(\{0.1,0.1,0.1,0.2,0.2,0.2,1\})$ and add Gaussian noises $\mathcal{N}\left(0, 0.02\right)$ to the cost observation $c_{i_t}^{(t)}$ of the subset $i_t$ if $c_{i_t} \geq 0.1$ . Using these sets of costs, the control sets are ordered such that $\mathcal{I}_i \subset \mathcal{I}_j \Rightarrow c_i < c_j$. These cost sets have fixed the optimal (i.e., last) control set to have a cost of 1. While these cost sets may (at first glance) seem arbitrary, it is the algorithms' relative performance across these cost sets rather than the absolute performance on a single cost set that allows us to understand the conditions under which particular algorithms perform better or worse. Real-world applications (unlike the experiments conducted here) will come with their own cost sets defined by real-world constraints. If the real costs can also be categorized in a similar relative way like the above cheap, moderate, and expensive cost sets, then the results are expected to be similar. Every probability distribution $\mathcal{P}_{\ell}$ is a truncated normal distribution with mean 0.5 and the same variance which is one of $0.02,0.04$ (the uniform distribution on $[0,1]$ has variance $1 / 12$ ).
We set the maximum budget for any experiment to 100 units. We compare the performance of our algorithm against that of the baseline Thompson sampling (TS-PSQ) and  UCB-PSQ developed in \cite{hayashi2022bayesian}. We evaluate a variant of UCB-CVS \cite{NEURIPS2023_090b23d5} called ETC-50, which involves 50 plays per group, with each group containing subsets having the same number of variables. This approach differs slightly from \cite{NEURIPS2023_090b23d5}, where groups are defined by subsets having the same cost due to the unknown cost setting. In the proposed method, we spend no more than 60 units of cost. The parameter $\alpha$ at the beginning of the exploitation phase is set to 0.1 and is halved after $d$ function evaluations.

\begin{figure*}
    \centering
    \includegraphics[width=0.45\linewidth]{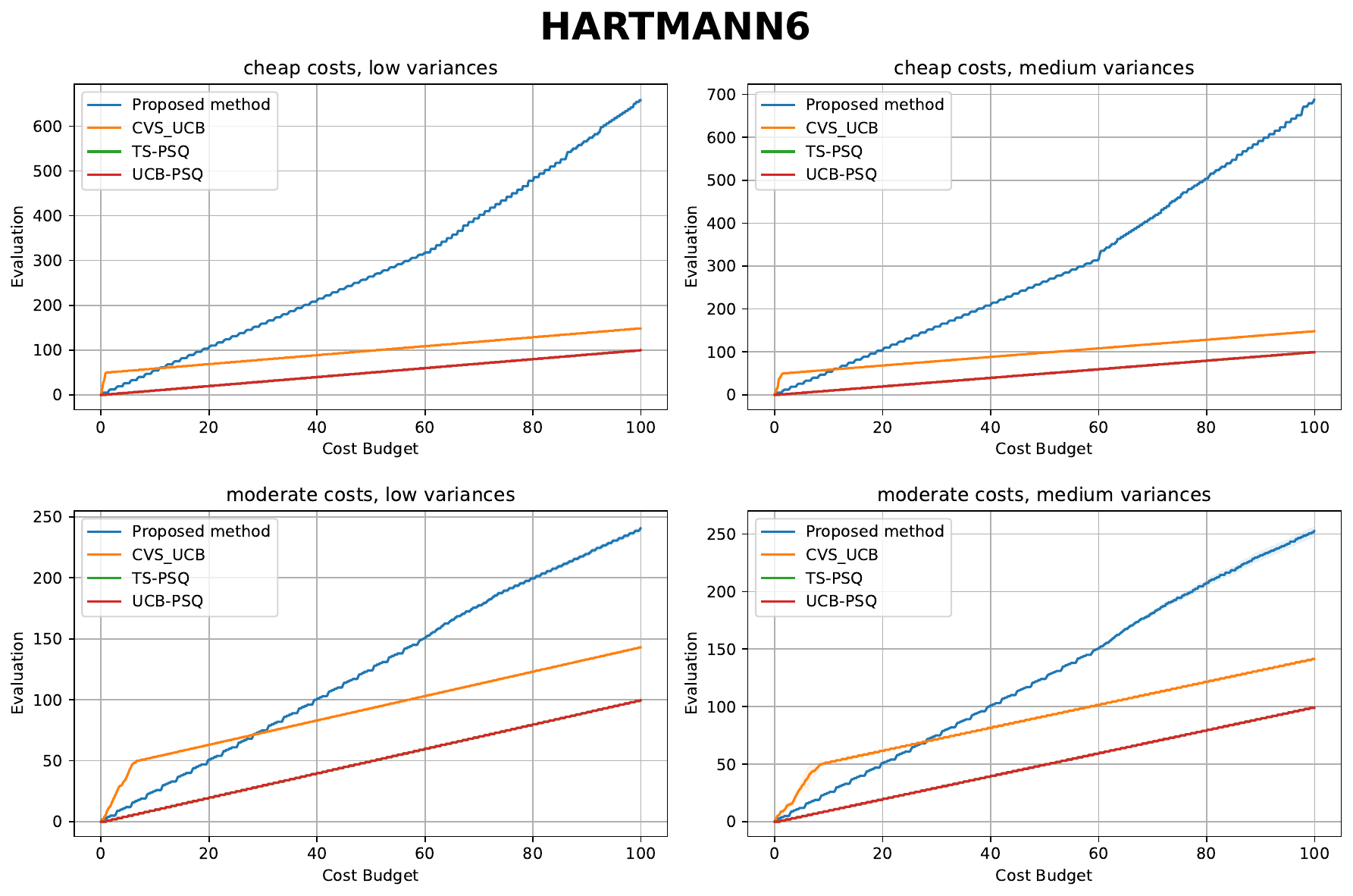}
    \includegraphics[width=0.45\linewidth]{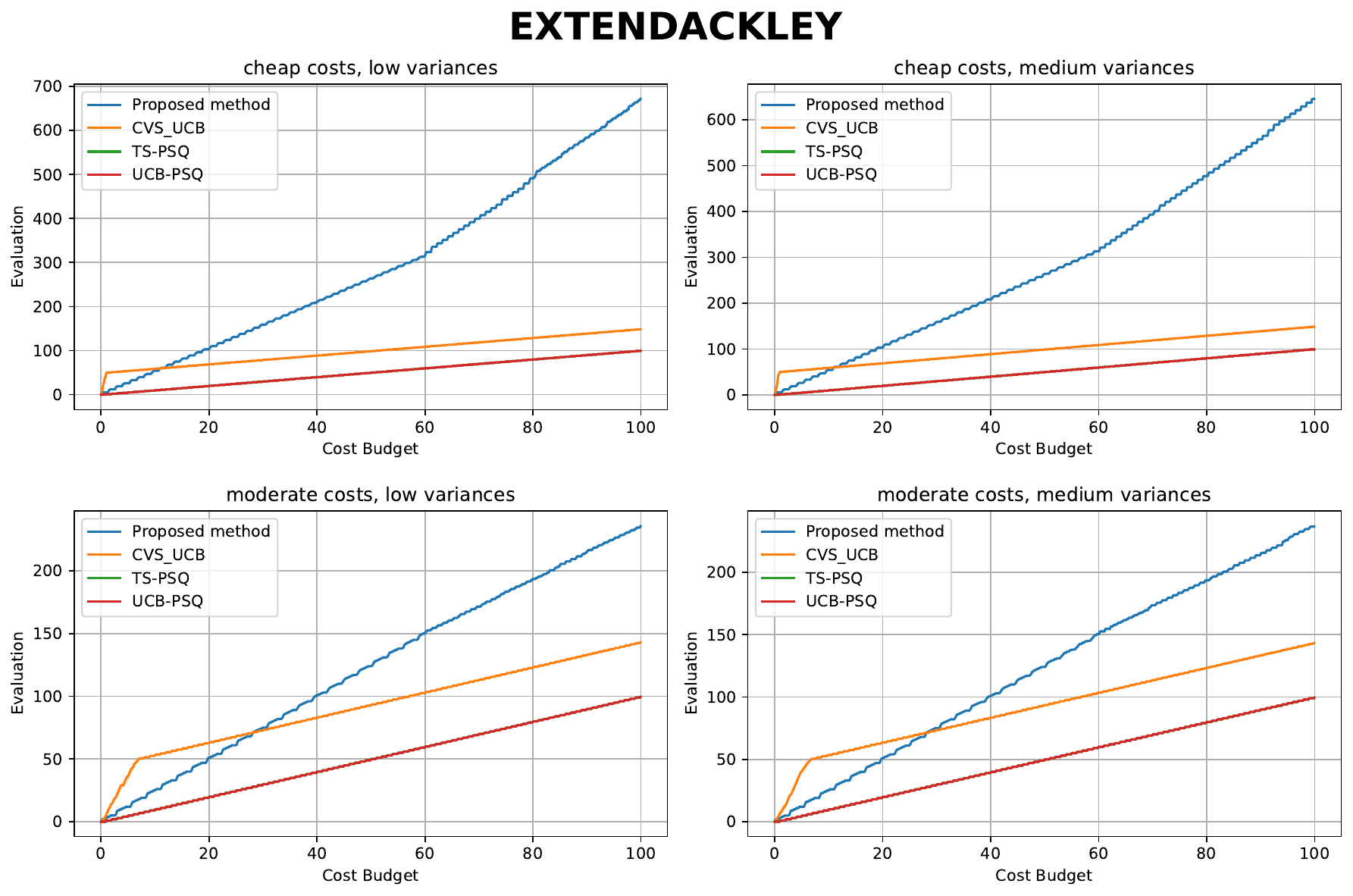}
    \caption{Number of evaluation against cost spent.}
    \label{fig:evalcomp}
\end{figure*}

Figure \ref{fig:main_results} shows the mean and standard error (over 5 seeds) of the simple regret $\min _{1 \leq t \leq \mathcal{T}(C)} \mathbb{E}\left[f\left(\left[\mathbf{x}^{i^+}, \mathbf{X}^{-i^+}\right]\right)\right]-\mathbb{E}\left[f\left(\left[\mathbf{x}^{i_t}, \mathbf{X}^{-i_t}\right]\right)\right]$ (lower is better) incurred against cost spent (budget) $C$ by each algorithm with varying objective functions, cost sets, and variances of distributions where $\mathcal{T}(C)$ denotes the maximum iteration reached after spending $C$. The simple regret encodes the value of the best solution an algorithm has chosen within a certain budget and is a measure of cost efficiency.
We observed that the naive methods are often less effective than the proposed method. The efficiency of the proposed algorithm stems from its strategy of prioritizing more plays on cost-effective feasible subsets, rather than solely targeting subsets likely to contain optimal solutions that are very expensive. Figure \ref{fig:evalcomp} compares the number of evaluations by the algorithms relative to the costs spent. The results indicate that the proposed algorithm conducts significantly more evaluations than the other algorithms once the exploitation phase starts. This implies that the baseline algorithms consistently evaluate the entire set, which is costly, rather than using the subset of valid variables, which is not only optimal but also less expensive. As a result, it leads to a lower number of evaluations compared to the proposed algorithm.

% \begin{figure}
%     \centering
%     \includegraphics[width=0.35\textwidth]{image/Main_hartmann6.pdf}
%     \caption{Comparison with the BO baselines on Hartmann}
%     \label{fig:hartmann6}
% \end{figure}

% \begin{figure}
%     \centering
%     \includegraphics[width=0.35\textwidth]{image/Main_extendackley.pdf}
%     \caption{Comparison with the BO baselines on Ackley}
%     \label{fig:ackley}
% \end{figure}

% \begin{figure}
%     \centering
%     \includegraphics[width=0.35\textwidth]{image/Main_extendlevy.pdf}
%     \caption{Comparison with the BO baselines on Levy}
%     \label{fig:levy}
% \end{figure}
\section{Related work}

To the best of our knowledge, UCB-CVS \cite{NEURIPS2023_090b23d5} is the only work that addresses the BOCVS setting at the time of writing. UCB-CVS uses an $\epsilon$-schedule as a relaxation technique, allowing cheaper control query sets to be selected for exploitation. The final control set is then chosen from this relaxed set by comparing their known costs. While UCB-CVS achieves a sublinear rate for cost-varying cumulative regret, it does not guarantee the minimization of the total incurred cost. Additionally, the practicality of the $\epsilon$-schedule is questionable due to the difficulty in selecting an appropriate value. Our proposed method addresses these issues, even in settings with unknown costs, making it more applicable to real-world problems.

The BOCVS setting can be viewed as a specific instance of the Causal Bayesian Optimization (CBO) framework \cite{AgliettiLPG20}, where the causal relationships among input variables are incorporated into the objective functions. In CBO, a subset of variables is selected for intervention, and their values are set to achieve optimal results while minimizing the overall intervention cost. Depending on the causal structure, intervening on certain variables may lead to a propagation of effects in the causal graph, affecting both the outcomes and associated costs. Therefore, the causal relationships in CBO can be seen as a generalized representation of input distribution in the BOCVS setting. CBO is a challenging problem, and to the best of our knowledge, most previous works \cite{AgliettiDGD21,AgliettiMKC23,GultchinABC23,BranchiniADD23} have not succeeded in establishing useful theoretical regret bounds. The only work addressing regret bounds in CBO is \citet{SussexM023}, which introduces such bounds under additional assumptions. However, their theoretical analysis does not account for minimizing the total incurred cost.

In BO field, there are also some settings where the learner is not only concerned about optimizing the objective function but also considers the trade-offs associated with multiple objectives that are intrinsic to several practical applications. \citet{HayashiHK22} introduced the setting \textit{with partially specified queries} (BOPSQ) where the learner simultaneously selects a control set and specifies their values to maximize the objective function. Hence, the learner is required to consider the extent to which the query variables contribute to the function’s output as well as which values the unspecified input variables will take. The key distinction between BOPSQ and the BOCVS setting is that BOPSQ does not incorporate cost constraints, which can significantly influence the learner’s decisions. On the other hand, multi-fidelity Bayesian Optimization (MFBO) \cite{Kandasamy16,PoloczekWF17,Takeno20} addresses scenarios where inexpensive, low-fidelity approximations of the true objective function are available. Each low-fidelity approximation is associated with a different level of model discrepancy and querying cost, with larger discrepancies generally resulting in lower costs. The goal of MFBO is to achieve an optimal objective value while minimizing the total cost of queries. Although the principle of paying less for potentially less informative queries is conceptually similar, the BOCVS setting differs in that the uncertainty arises from the randomness of inputs not included in the control set, rather than from querying a lower-cost approximation as in MFBO.

BOCVS can also be viewed as a specialized version of multi-armed bandits (MAB) with cost constraints \cite{CayciE020,DingQZL13,SinhaSKA21}. In this context, each control set functions as an arm, and pulling an arm yields a reward along with an associated cost. The goal of the algorithm is to select a sequence of arms to maximize the expected total reward while ensuring that the costs of pulling these arms stay within a budget constraint. However, unlike in the standard cost-constrained MAB setting, BOCVS requires the learner not only to pull an arm but also to specify parameters, such as the values for the control set associated with that arm. This simultaneous decision-making process adds a layer of complexity, making BOCVS more challenging than traditional cost-constrained MAB problems.

\section{Conclusion}
We introduce an extension of the BOCVS problem where the costs of subsets are random and unknown, and we develop a cost-sensitive algorithm with theoretical guarantees, demonstrating sub-linear cumulative regret for both objective function value and cost. Our empirical evaluation compared the proposed algorithm's performance against baseline methods across various experimental conditions, including both synthetic and real-world datasets.

\bibliography{ref}

\begin{thebibliography}{25}
\providecommand{\natexlab}[1]{#1}

\bibitem[{Aglietti et~al.(2021)Aglietti, Dhir, Gonz{\'{a}}lez, and Damoulas}]{AgliettiDGD21}
Aglietti, V.; Dhir, N.; Gonz{\'{a}}lez, J.; and Damoulas, T. 2021.
\newblock Dynamic Causal Bayesian Optimization.
\newblock In Ranzato, M.; Beygelzimer, A.; Dauphin, Y.~N.; Liang, P.; and Vaughan, J.~W., eds., \emph{Advances in Neural Information Processing Systems 34: Annual Conference on Neural Information Processing Systems 2021, NeurIPS 2021, December 6-14, 2021, virtual}, 10549--10560.

\bibitem[{Aglietti et~al.(2020)Aglietti, Lu, Paleyes, and Gonz{\'{a}}lez}]{AgliettiLPG20}
Aglietti, V.; Lu, X.; Paleyes, A.; and Gonz{\'{a}}lez, J. 2020.
\newblock Causal Bayesian Optimization.
\newblock In Chiappa, S.; and Calandra, R., eds., \emph{The 23rd International Conference on Artificial Intelligence and Statistics, {AISTATS} 2020, 26-28 August 2020, Online [Palermo, Sicily, Italy]}, volume 108 of \emph{Proceedings of Machine Learning Research}, 3155--3164. {PMLR}.

\bibitem[{Aglietti et~al.(2023)Aglietti, Malek, Ktena, and Chiappa}]{AgliettiMKC23}
Aglietti, V.; Malek, A.; Ktena, I.; and Chiappa, S. 2023.
\newblock Constrained Causal Bayesian Optimization.
\newblock In Krause, A.; Brunskill, E.; Cho, K.; Engelhardt, B.; Sabato, S.; and Scarlett, J., eds., \emph{International Conference on Machine Learning, {ICML} 2023, 23-29 July 2023, Honolulu, Hawaii, {USA}}, volume 202 of \emph{Proceedings of Machine Learning Research}, 304--321. {PMLR}.

\bibitem[{Branchini et~al.(2023)Branchini, Aglietti, Dhir, and Damoulas}]{BranchiniADD23}
Branchini, N.; Aglietti, V.; Dhir, N.; and Damoulas, T. 2023.
\newblock Causal Entropy Optimization.
\newblock In Ruiz, F. J.~R.; Dy, J.~G.; and van~de Meent, J., eds., \emph{International Conference on Artificial Intelligence and Statistics, 25-27 April 2023, Palau de Congressos, Valencia, Spain}, volume 206 of \emph{Proceedings of Machine Learning Research}, 8586--8605. {PMLR}.

\bibitem[{Cayci, Eryilmaz, and Srikant(2020)}]{CayciE020}
Cayci, S.; Eryilmaz, A.; and Srikant, R. 2020.
\newblock Budget-Constrained Bandits over General Cost and Reward Distributions.
\newblock In Chiappa, S.; and Calandra, R., eds., \emph{The 23rd International Conference on Artificial Intelligence and Statistics, {AISTATS} 2020, 26-28 August 2020, Online [Palermo, Sicily, Italy]}, volume 108 of \emph{Proceedings of Machine Learning Research}, 4388--4398. {PMLR}.

\bibitem[{Chowdhury and Gopalan(2017)}]{chowdhury2017kernelized}
Chowdhury, S.~R.; and Gopalan, A. 2017.
\newblock On kernelized multi-armed bandits.
\newblock In \emph{International Conference on Machine Learning}, 844--853. PMLR.

\bibitem[{Ding et~al.(2013)Ding, Qin, Zhang, and Liu}]{DingQZL13}
Ding, W.; Qin, T.; Zhang, X.; and Liu, T. 2013.
\newblock Multi-Armed Bandit with Budget Constraint and Variable Costs.
\newblock In desJardins, M.; and Littman, M.~L., eds., \emph{Proceedings of the Twenty-Seventh {AAAI} Conference on Artificial Intelligence, July 14-18, 2013, Bellevue, Washington, {USA}}, 232--238. {AAAI} Press.

\bibitem[{Dua, Graff et~al.(2017)}]{dua2017uci}
Dua, D.; Graff, C.; et~al. 2017.
\newblock UCI machine learning repository, 2017.
\newblock \emph{URL http://archive. ics. uci. edu/ml}, 7(1): 62.

\bibitem[{Gultchin et~al.(2023)Gultchin, Aglietti, Bellot, and Chiappa}]{GultchinABC23}
Gultchin, L.; Aglietti, V.; Bellot, A.; and Chiappa, S. 2023.
\newblock Functional causal Bayesian optimization.
\newblock In Evans, R.~J.; and Shpitser, I., eds., \emph{Uncertainty in Artificial Intelligence, {UAI} 2023, July 31 - 4 August 2023, Pittsburgh, PA, {USA}}, volume 216 of \emph{Proceedings of Machine Learning Research}, 756--765. {PMLR}.

\bibitem[{Hayashi, Honda, and Kashima(2022{\natexlab{a}})}]{hayashi2022bayesian}
Hayashi, S.; Honda, J.; and Kashima, H. 2022{\natexlab{a}}.
\newblock Bayesian optimization with partially specified queries.
\newblock \emph{Machine Learning}, 1--30.

\bibitem[{Hayashi, Honda, and Kashima(2022{\natexlab{b}})}]{HayashiHK22}
Hayashi, S.; Honda, J.; and Kashima, H. 2022{\natexlab{b}}.
\newblock Bayesian optimization with partially specified queries.
\newblock \emph{Mach. Learn.}, 111(3): 1019--1048.

\bibitem[{Kandasamy et~al.(2016)Kandasamy, Dasarathy, Oliva, Schneider, and P{\'{o}}czos}]{Kandasamy16}
Kandasamy, K.; Dasarathy, G.; Oliva, J.~B.; Schneider, J.~G.; and P{\'{o}}czos, B. 2016.
\newblock Gaussian Process Bandit Optimisation with Multi-fidelity Evaluations.
\newblock In Lee, D.~D.; Sugiyama, M.; von Luxburg, U.; Guyon, I.; and Garnett, R., eds., \emph{Advances in Neural Information Processing Systems 29: Annual Conference on Neural Information Processing Systems 2016, December 5-10, 2016, Barcelona, Spain}, 992--1000.

\bibitem[{Kirschner and Krause(2018)}]{kirschner2018information}
Kirschner, J.; and Krause, A. 2018.
\newblock Information directed sampling and bandits with heteroscedastic noise.
\newblock In \emph{Conference On Learning Theory}, 358--384. PMLR.

\bibitem[{Kirschner et~al.(2022)Kirschner, Mutn\'y, Krause, Coello~de Portugal, Hiller, and Snuverink}]{Kirschner2022}
Kirschner, J.; Mutn\'y, M.; Krause, A.; Coello~de Portugal, J.; Hiller, N.; and Snuverink, J. 2022.
\newblock Tuning particle accelerators with safety constraints using Bayesian optimization.
\newblock \emph{Phys. Rev. Accel. Beams}, 25: 062802.

\bibitem[{Kruger et~al.(2011)Kruger, Shih, Hattingh, and {van Niekerk}}]{KRUGER2011783}
Kruger, G.~H.; Shih, A.~J.; Hattingh, D.~G.; and {van Niekerk}, T.~I. 2011.
\newblock Intelligent machine agent architecture for adaptive control optimization of manufacturing processes.
\newblock \emph{Advanced Engineering Informatics}, 25(4): 783--796.
\newblock Special Section: Advances and Challenges in Computing in Civil and Building Engineering.

\bibitem[{Poloczek, Wang, and Frazier(2017)}]{PoloczekWF17}
Poloczek, M.; Wang, J.; and Frazier, P.~I. 2017.
\newblock Multi-Information Source Optimization.
\newblock In Guyon, I.; von Luxburg, U.; Bengio, S.; Wallach, H.~M.; Fergus, R.; Vishwanathan, S. V.~N.; and Garnett, R., eds., \emph{Advances in Neural Information Processing Systems 30: Annual Conference on Neural Information Processing Systems 2017, December 4-9, 2017, Long Beach, CA, {USA}}, 4288--4298.

\bibitem[{Shields et~al.(2021)Shields, Stevens, Li, Parasram, Damani, Alvarado, Janey, Adams, and Doyle}]{ShieldsSLPDAJAD21}
Shields, B.~J.; Stevens, J.~M.; Li, J.; Parasram, M.; Damani, F.~N.; Alvarado, J. I.~M.; Janey, J.~M.; Adams, R.~P.; and Doyle, A.~G. 2021.
\newblock Bayesian reaction optimization as a tool for chemical synthesis.
\newblock \emph{Nat.}, 590(7844): 89--96.

\bibitem[{Sinha et~al.(2021)Sinha, Sankararaman, Kazerouni, and Avadhanula}]{SinhaSKA21}
Sinha, D.; Sankararaman, K.~A.; Kazerouni, A.; and Avadhanula, V. 2021.
\newblock Multi-Armed Bandits with Cost Subsidy.
\newblock In Banerjee, A.; and Fukumizu, K., eds., \emph{The 24th International Conference on Artificial Intelligence and Statistics, {AISTATS} 2021, April 13-15, 2021, Virtual Event}, volume 130 of \emph{Proceedings of Machine Learning Research}, 3016--3024. {PMLR}.

\bibitem[{Slivkins et~al.(2019)}]{slivkins2019introduction}
Slivkins, A.; et~al. 2019.
\newblock Introduction to multi-armed bandits.
\newblock \emph{Foundations and Trends{\textregistered} in Machine Learning}, 12(1-2): 1--286.

\bibitem[{Snoek, Larochelle, and Adams(2012)}]{Snoek2012}
Snoek, J.; Larochelle, H.; and Adams, R.~P. 2012.
\newblock Practical Bayesian Optimization of Machine Learning Algorithms.
\newblock In Pereira, F.; Burges, C.; Bottou, L.; and Weinberger, K., eds., \emph{Advances in Neural Information Processing Systems}, volume~25. Curran Associates, Inc.

\bibitem[{Srinivas et~al.(2009)Srinivas, Krause, Kakade, and Seeger}]{srinivas2009gaussian}
Srinivas, N.; Krause, A.; Kakade, S.~M.; and Seeger, M. 2009.
\newblock Gaussian process optimization in the bandit setting: No regret and experimental design.
\newblock \emph{arXiv preprint arXiv:0912.3995}.

\bibitem[{Sussex, Makarova, and Krause(2023)}]{SussexM023}
Sussex, S.; Makarova, A.; and Krause, A. 2023.
\newblock Model-based Causal Bayesian Optimization.
\newblock In \emph{The Eleventh International Conference on Learning Representations, {ICLR} 2023, Kigali, Rwanda, May 1-5, 2023}. OpenReview.net.

\bibitem[{Takeno et~al.(2020)Takeno, Fukuoka, Tsukada, Koyama, Shiga, Takeuchi, and Karasuyama}]{Takeno20}
Takeno, S.; Fukuoka, H.; Tsukada, Y.; Koyama, T.; Shiga, M.; Takeuchi, I.; and Karasuyama, M. 2020.
\newblock Multi-fidelity Bayesian Optimization with Max-value Entropy Search and its Parallelization.
\newblock In \emph{Proceedings of the 37th International Conference on Machine Learning, {ICML} 2020, 13-18 July 2020, Virtual Event}, volume 119 of \emph{Proceedings of Machine Learning Research}, 9334--9345. {PMLR}.

\bibitem[{Tay et~al.(2023)Tay, Foo, Urano, Leong, and Low}]{NEURIPS2023_090b23d5}
Tay, S.; Foo, C.~S.; Urano, D.; Leong, R.; and Low, B. K.~H. 2023.
\newblock Bayesian Optimization with Cost-varying Variable Subsets.
\newblock In Oh, A.; Naumann, T.; Globerson, A.; Saenko, K.; Hardt, M.; and Levine, S., eds., \emph{Advances in Neural Information Processing Systems}, volume~36, 3008--3031. Curran Associates, Inc.

\bibitem[{Williams and Rasmussen(2006)}]{williams2006gaussian}
Williams, C.~K.; and Rasmussen, C.~E. 2006.
\newblock \emph{Gaussian processes for machine learning}, volume~2.
\newblock MIT press Cambridge, MA.

\end{thebibliography}

\appendix
\onecolumn
\setcounter{section}{1}
\section{Theoretical Proof}
\subsection{Lemmas}
\begin{lemma}
\label{lemma1}
     \cite{chowdhury2017kernelized} Let $\left(\mathrm{x}_t\right)_{t=1}^T$ be a sequence of queries selected by some algorithm. Then, the mutual information $I\left(\mathbf{y}_{1: T} ; \mathbf{f}_{1: T}\right)$ between the noisy observations $\mathbf{y}_{1: T}$ and the function values $\mathbf{f}_{1: T}$ at the queries is given by
$$
\begin{aligned}
    I\left(\mathbf{y}_{1: T} ; \mathbf{f}_{1: T}\right)&=\frac{1}{2} \log \left|\mathbf{I}+\lambda^{-1} \mathbf{K}_t\right| \\
    &=\frac{1}{2} \sum_{t=1}^T \log \left(1+\lambda^{-1} \sigma_{t-1}^2\left(\mathbf{x}_t\right)\right)
\end{aligned}
$$
\end{lemma}

\begin{lemma}
\label{lemma2}
    \cite{chowdhury2017kernelized} Let $\beta_t:=B+\sigma \sqrt{2\left(\gamma_{t-1}(\mathcal{X})+1+\log (1 / \delta)\right)}$ where $B$ is the upper bound of the RKHS norm of $f$. With probability at least $1-\delta$, for all $\mathbf{x} \in \mathcal{X}$ and $t \geq 1$,
$$
\left|\mu_{t-1}(\mathbf{x})-f(\mathbf{x})\right| \leq \beta_t \sigma_{t-1}(\mathbf{x})
$$
where $\mu_{t-1}$ and $\sigma_{t-1}$ are defined in (1) with $\lambda=1+\eta$ and $\eta:=2 / T$.
\end{lemma} 

\begin{lemma}
    \cite{chowdhury2017kernelized} Let $\left(\mathbf{x}_t\right)_{t=1}^T$ be a sequence of queries selected by some algorithm. Then
$$
\sum_{t=1}^T \sigma_{t-1}\left(\mathbf{x}_t\right) \leq \sqrt{4(T+2) \gamma_T(\mathcal{X})} .
$$
\label{lemma3}
\end{lemma}

\begin{lemma}
\cite{kirschner2018information}
     Let $X_t$ be any non-negative stochastic process adapted to a filtration $\left\{\mathcal{F}_t\right\}$, and define $m_t:=\mathbb{E}\left[X_t \mid \mathcal{F}_{t-1}\right]$. Further assume that $X_t \leq b_t$ for a fixed, non-decreasing sequence $\left(b_t\right)_{t \geq 1}$. If $b_T \geq 1$, with probability at least $1-\delta$, for any $T \geq 1$,
$$
\sum_{t=1}^T m_t \leq 2 \sum_{t=1}^T X_t+4 b_T \log \frac{m}{\delta}+8 b_T \log \left(4 b_T\right)+1
$$
\label{lemma4}
\end{lemma}

\subsection{Proof of Theorem \ref{theo:1}}
We have that 
$$
\begin{aligned}
& R_T^f=\sum_{i \in[m]: i \neq i^{+}} \sum_{k=0}^{\tau-1}(1-\alpha) E\left[f\left(x^{i^{+}} ; X^{-i^{+}}\right)\right]-E\left[f\left(x_{k m+i}^i ; X^{-i}\right)\right] \\
& +\sum_{k=0}^{\tau-1}(1-\alpha) E\left[f\left(x^{i^{+}} ; X^{-i^{+}}\right)\right]-E\left[f\left(x_{k m+i^{+}}^{i^{+}} ; X^{-i^{+}}\right)\right] \\
& +\sum_{t=m \tau+1}^T(1-\alpha) E\left[f\left(x^{i^{+}} ; X^{-i^{+}}\right)\right]-E\left[f\left(x_t^{i_t} ; X^{-i_t}\right)\right] \\
& R_T^c=\sum_{t=1}^{m \tau} \max \left\{\left(c_{i_t}-c_{i^*}\right) ; 0\right\}+\sum_{t=m \tau+1}^T \max \left\{\left(c_{i_t}-c_{i^*}\right) ; 0\right\}
\end{aligned}
$$
\textbf{Phase 1:} Using Lemma \ref{lemma1},\ref{lemma2},\ref{lemma3},\ref{lemma4}, with probability at least $1-2\delta$, we have:
$$
\begin{aligned}
& \sum_{i \in[m] ; i \neq i^{+}} \sum_{k=0}^{\tau-1}(1-\alpha) E\left[f\left(x^{i^{+}} ; X^{-i^{+}}\right)\right]-E\left[f\left(x_{k m+i}^i ; X^{-i}\right)\right] \\
&\leq 2(m-1) \tau M \\
\text{and} \\
& \sum_{k=0}^{\tau-1}(1-\alpha) E\left[f\left(\mathrm{x}^{i^{+}} ; X^{-i^{+}}\right)\right]-E\left[f\left(x_{k m+i^{+}}^{i^{+}} ; X^{-i^{+}}\right)\right] \\
\leq & \sum_{k=0}^{\tau-1}(1-\alpha) E\left[u_{k m+i^{+}-1}\left(\mathrm{x}^{i^{+}} ; X^{-i^{+}}\right)\right]-E\left[f\left(\mathrm{x}_{k m+i^{+}}^{i^{+}} ; X^{-^{+}}\right)\right] \\
\leq & \sum_{k=0}^{\tau-1}(1-\alpha) E\left[u_{k m+i^{+}-1}\left(\mathrm{x}_{k m+i^{+}} ; X^{-i^{+}}\right)\right]-E\left[f\left(\mathrm{x}_{k m+i^{+}}{ }^{+} ; X^{-i^{+}}\right)\right] \\
\leq & \sum_{k=0}^{\tau-1}(1-\alpha) E\left[u_{k m+i^{+}-1}\left(\mathrm{x}_{k m+i^{+}} ; X^{-i^{+}}\right)\right]-E\left[l_{k m+i^{+}-1}\left(\mathrm{x}_{k m+i^{+}}^{i^{+}} ; X^{-i^{+}}\right)\right] \\
\leq & \sum_{k=0}^{\tau-1} 2 \beta_{k m+i^{+}} E\left[\sigma_{k m+i^{+}-1}\left(\mathrm{x}_{k m+i^{+}}{ }^{+} ; X^{-i^{+}}\right)\right] \\
\leq & 2 \beta_{\tau m}\left(2 \sum_{k=0}^{\tau-1} \sigma_{k m+i^{+}-1}\left(\mathrm{x}_{k m+i^{+}}\right)+4 \log \frac{m}{\delta}+8 \log 4+1\right) \\
\leq & 2 \beta_{\tau m}\left(2 \sqrt{4(\tau+2) \gamma_\tau(x)}+4 \log \frac{m}{\delta}+8 \log 4+1\right)
\end{aligned}
$$
\textbf{Phase 2:} For every $t>0$ with a probability of at least $1-\delta$, we have for every $i_0 \in C^*$ :
$$
\begin{aligned}
& (1-\alpha) \max _{\left(i ; x^i\right) \in[m] \times x^i} E\left[l_{t-1}\left(x^i ; X^{-i}\right)\right] \\
& \leq(1-\alpha) E\left[f\left(x^{i^{+}} ; X^{-i^{+}}\right)\right] \\
& \leq \max _{x^{i_0} \in X^{i_0}} E\left[f\left(x^{i_0} ; X^{-i_0}\right)\right] \\
& \leq \max _{x^{i_0} \in X^{i_0}} E\left[u_{t-1}\left(x^{i_0} ; X^{-i_0}\right)\right]
\end{aligned}
$$
Therefore, for every $t>m\tau$ with a probability of at least $1-\delta, C^* \in S_1 \neq \emptyset$. Let $k_0<\tau$ satisfy:
$$
k_0=\arg \min _{k<\tau} \beta_{k m+i^{+}} \mathbb{E}\left(\left[\sigma_{k m+i^{+}-1}\left[\mathbf{x}_{k m+i^{+}}^{i^{+}} ; \mathbf{X}^{-i^{+}}\right]\right)\right]
$$
From here we have:
$$
\begin{aligned}
& \beta_{k_0 m+i^{+}} \mathbb{E}\left(\left[\sigma_{k_0 m+i^{+}-1}\left[\mathbf{x}_{k m+i^{+}}^{i^{+}} ; \mathbf{X}^{-i^{+}}\right]\right)\right] \\
& \leq \frac{1}{\tau} \sum_{k=0}^{\tau-1} \beta_{k m+i^{+}} \mathbb{E}\left(\left[\sigma_{k m+i^{+}-1}\left[\mathbf{x}_{k m+i^{+}}^{i^{+}} ; \mathbf{X}^{-i^{+}}\right]\right)\right] \\
& \leq \frac{1}{\tau} \beta_{\tau m}\left(2 \sqrt{4(\tau+2) \gamma_\tau(\mathcal{X})}+4 \log \frac{m}{\delta}+8 \log 4+1\right)
\end{aligned}
$$
Thus, with a probability of $1-\delta$, for all $t>m \tau \geq m k_0+i+$ :
$$
\overline{lcb}_{t-1}> E\left[l_{m k_0+i^{+}-1}\left(x_{m k_0+i^{+}}^{i^{+}} ; X^{-i^{+}}\right)\right]
$$
Using Lemma \ref{lemma1},\ref{lemma2},\ref{lemma3},\ref{lemma4}, with probability at least $1-2\delta$, we have:
$$
\begin{aligned}
& \sum_{t=m \tau+1}^T(1-\alpha) E\left[f\left(x^{i^{+}} ; X^{-i^{+}}\right)\right]-E\left[f\left(x_t^{i_t} ; X^{-i_t}\right)\right] \\
& \leq \sum_{\substack{t=m \tau+1 \\
T}}^T(1-\alpha) E\left[u_{m k_0+i^{+}-1}\left(x_{m k_0+i^{+}}^{i^{+}} ; X^{-i^{+}}\right)\right] -E\left[f\left(x_t^{i_t} ; X^{-i_t}\right)\right] \\
& \leq \sum_{t=m \tau+1}^T(1-\alpha) E\left[l_{m k_0+i^{+}-1}\left(x_{m k_0+i^{+}}^{i^{+}} ; X^{-i^{+}}\right)\right] \\
& +2(1-\alpha) \beta_{m k_0+i^{+}} E\left[\sigma_{m k_0+i^{+}-1}\left(x_{m k_0+i^{+}}^{i^{+}} ; X^{-i^{+}}\right)\right]-E\left[f\left(x_t^{i_t} ; X^{-i_t}\right)\right] \\
& \leq \sum_{t=m \tau+1}^T E\left[u_{t-1}\left(x^{i_t} ; X^{-i_t}\right)\right]-E\left[f\left(x_t^{i_t} ; X^{-i_t}\right)\right] \\
&+2 \frac{T-m \tau}{\tau} \beta_{\tau m}\left(2 \sqrt{4(\tau+2) \gamma_\tau(X)}+4 \log \frac{m}{\delta}+8 \log 4+1\right) \\
& \leq 2 \sum_{t=m \tau+1}^T \beta_t E\left[\sigma_{t-1}\left(x^{i_t} ; X^{-i_t}\right)\right]\\
&+2 \frac{T-m \tau}{\tau} \beta_{\tau m}\left(2 \sqrt{4(\tau+2) \gamma_\tau(\mathcal{X})}+4 \log \frac{m}{\delta}+8 \log 4+1\right) \\
& \leq 2 \beta_T \sum_{i=1}^m \sum_{t \in \tilde{T}_i ; t>m \tau} \mathbb{E}\left[\sigma_{t-1}\left(\left[\mathrm{x}^{i_t} ; \mathbf{X}^{-i_t}\right]\right)\right] \\
&+2 \frac{T-m \tau}{\tau} \beta_{\tau m}\left(2 \sqrt{4(\tau+2) \gamma_\tau(\mathcal{X})}+4 \log \frac{m}{\delta}+8 \log 4+1\right) \\
& \leq 2 \beta_T \sum_{i=1}^m\left(2 \sum_{t \in \tilde{T}_i i t>m \tau} \sigma_{t-1}\left(\mathrm{x}_t\right)+4 \log \frac{m}{\delta}+8 \log 4+1\right) \\
&+2 \frac{T-m \tau}{\tau} \beta_{\tau m}\left(2 \sqrt{4(\tau+2) \gamma_\tau(\mathcal{X})}+4 \log \frac{m}{\delta}+8 \log 4+1\right) \\
& \leq 2 \beta_T\left(2 \sqrt{4(T+2-m \tau) \gamma_{T-m \tau}(\mathcal{X})}
+m\left(4 \log \frac{m}{\delta}+8 \log (4)+1\right)\right) \\
&+2 \frac{T-m \tau}{\tau} \beta_{\tau m}\left(2 \sqrt{4(\tau+2) \gamma_\tau(\mathcal{X})}+4 \log \frac{m}{\delta}+8 \log 4+1\right)
\end{aligned}
$$
With probability at least $1-\delta$; we have the optimal subset $i^* \in C_* \in \mathcal{S}_1 ; \forall t>m \tau$. Using union bound and Lemma 1.6 in \cite{slivkins2019introduction}; with probability at least $1-\delta-2 / T^2$; we have
$$
\begin{aligned}
&r_t^c=\max \left\{\left(c_{i_t}-c_{i^*}\right) ; 0\right\} \leq 2 \beta_{i_t}(t)=2 \sqrt{\frac{2 \log T}{T_{i_t}(t)}} \\
&\sum_{t=m \tau+1}^T r_t^c \leq \sum_{t=m \tau+1}^T 2 \sqrt{\frac{2 \log T}{T_{i_t}(t)}} \\
&\leq 2 \sum_{i=1}^m\left(T_i-\tau\right) \sqrt{\frac{2 \log T}{\tau}}=2(T-m \tau) \sqrt{\frac{2 \log T}{\tau}}
\end{aligned}
$$
\subsection{Proof Lemma \ref{lm:2}}
With at least $1-\delta$, for every $t>m \tau$, we have
$$
\begin{aligned}
& \overline{l c b}_{t-1} \geq E\left[l_{m k_0+i^{+}-1}\left(x_{m k_0+i^{+}}^{i^{+}}; X^{-i^{+}}\right)\right] \\
& =E\left[u_{m k_0+i^{+}-1}\left(x_{m k_0+i^{+}}^{i^{+}} ; X^{-i^{+}}\right)\right]-2 \beta_{m k_0+i^{+}} E\left[\sigma_{m k_0+i^{+}-1}\left(x_{m k_0+i^{+}}^{i^{+}} ; X^{-i^{+}}\right)\right] \\
& \geq E\left[u_{m k_0+i^{+}-1}\left(x^{i^{+}} ; X^{-i^{+}}\right)\right] -2 \beta_{m k_0+i^{+}} E\left[\sigma_{m k_0+i^{+}-1}\left(x_{\left.\left.m k_0+i^{+} ; X^{-i^{+}}\right)\right]}\right)\right] \\
& \geq E\left[f\left(x^{i^{+}} ; X^{-i^{+}}\right)\right]-\frac{1}{\tau} \beta_{\tau m}\left(2 \sqrt{4(\tau+2) \gamma_\tau(X)}+4 \log \frac{m}{\delta}+8 \log 4+1\right) \\
& \geq E\left[f\left(x^{i^{+}} ; X^{-i^{+}}\right)\right]-\frac{\tilde{\epsilon}}{2-\sigma}
\end{aligned}
$$
and for every $i^{\prime} \notin C_*$, we have that
$$
\begin{aligned}
& \overline{u c b}_{i, t-1} \leq E\left[u_{m k_i+i-1}\left(x_{m k_i+i}^i ; X^{-i}\right)\right] \\
& =E\left[l_{m k_i+i-1}\left(x_{m k_i+i}^i ; X^{-i}\right)\right]+2 \beta_{m k_0+i} E\left[\sigma_{m k_i+i-1}\left(x_{m k_0+i^{+}}^i ; X^{-i}\right)\right] \\
& \leq E\left[f\left(x_{m k_i+i}^i ; X^{-i}\right)\right]+2 \beta_{m k_0+i} E\left[\sigma_{m k_i+i-1}\left(x_{m k_0+i^{+}}^i ; X^{-i}\right)\right] \\
& \leq \max _{x^i \in X^i} E\left[f\left(x^i ; X^{-i}\right)\right]+\frac{1}{\tau} \beta_{\tau m}\left(2 \sqrt{4(\tau+2) \gamma_\tau(X)}+4 \log \frac{m}{\delta}+8 \log 4+1\right) \\
& \leq(1-\alpha) E\left[f\left(x^{i^{+}} ; X^{-i^{+}}\right)\right]-\tilde{\epsilon}+\frac{\tilde{\epsilon}}{2-\sigma}
\end{aligned}
$$
where
$$
k_i=\arg \min _{k<\tau} \beta_{k m+i} \mathbb{E}\left(\left[\sigma_{k m+\mathrm{i}-1}\left[\mathbf{x}_{k m+i^{+}}^i ; \mathbf{X}^{-\mathrm{i}}\right]\right)\right]
$$
\quad \quad \quad Then it implies that for every $i^{\prime} \notin C_*$, we have
$$
(1-\alpha) \overline{l c b}_{t-1} \geq \overline{u c b}_{i, t-1}
$$
$\operatorname{Or} C_*=S_1$

\subsection{Proof of Theorem \ref{theo:2}}
Phase 1:
$$
\sum_{i \in[m]} \sum_{k=0}^{\tau-1}(1-\alpha) E\left[f\left(x^{i^{+}} ; X^{-i^{+}}\right)\right]-E\left[f\left(x_{k m+i}^i ; X^{-i}\right)\right] \leq 2 \tau M
$$

Phase 2:
Using Lemma \ref{lm:2}, we have $S_1=C_*$ for every $t>m \tau$. It implies that $i_t \in C_*$, then we have:
$$
(1-\alpha) E\left[f\left(x^{i^{+}} ; X^{-i^{+}}\right)\right] \leq E\left[u_{t-1}\left(x^{i_t} ; X^{-i_t}\right)\right]
$$

We also have
$$
\begin{aligned}
& \sum_{t=m \tau+1}^T(1-\alpha) E\left[f\left(x^{i^{+}} ; X^{-i^{+}}\right)\right]-E\left[f\left(x_t^{i_t} ; X^{-i_t}\right)\right] \\
\leq & \sum_{t=m \tau+1}^T E\left[u_{t-1}\left(x^{i_t} ; X^{-i_t}\right)\right]-E\left[f\left(x_t^{i_t} ; X^{-i_t}\right)\right] \\
\leq & 2 \sum_{t=m \tau+1}^T \beta_t E\left[\sigma_{t-1}\left(x^{i_t} ; X^{-i_t}\right)\right] \\
\leq & 2 \beta_T\left(2 \sqrt{4(T+2-m \tau) \gamma_{T-m \tau}(X)}+m\left(4 \log \frac{1}{\delta}+8 \log (4)+1\right)\right)
\end{aligned}
$$
$R_T^c$ is the same as in Theorem \ref{theo:1}.

\section{Additional details on the implementation and the empirical evaluation}
\subsection{Control sets}
The control sets described here are given in an order corresponding to their costs given in the Experiment section.
\paragraph{Hartmann and Levy (12-D, 6-d)}: We use control sets $\{\{1,2,3\},\{4,5,6\},\{7,8,9\},\{10,11,12\},\{1,2,3,4,5,6\},$ $\{7,8,9,10,11,12\}\{1,2,3,4,5,6,7,8,9,10,11,12\}\}$.

\paragraph{Plant growth simulator (5-D)}: The plant growth simulator is a GP built from private data collected on the maximum leaf area achieved by Marchantia plants depending on input variables $\mathrm{Ca}, \mathrm{B}, \mathrm{NH}_3, \mathrm{~K}$, and pH. We use min-max feature scaling to scale all input variables to $[0,1]$ and standardize the output values. We use the posterior mean of the GP as the objective function. We use control sets $\{\{1,2\},\{3,4\},\{4,5\},\{1,2,3\},\{2,3,4\},\{3,4,5\},\{1,2,3,4,5\}\}$.

\paragraph{Airfoil self-noise (5-D)}: We use the airfoil self-noise dataset from \cite{dua2017uci}. To scale all input variables to $[0,1]$, we first take the natural logarithm of variables 1 and 5 , then do min-max feature scaling on all input variables. We also standardize the output values. We then feed the data into a default SingleTaskGP from BoTorch and use the posterior mean as the objective function. We use control sets $\{\{4,5\},\{2,5\},\{1,4\},\{2,3\},\{3,5\},\{1,2\},\{3,4\}\}$.
\subsection{Baselines}
For UCB-PSQ, TS-PSQ and UCB-CVS, we  use implementation from \cite{NEURIPS2023_090b23d5} (\url{https://github.com/sebtsh/cost-varying-variable-subsets-BO}). For the experiments, we set $\beta_t=2$ for all $t$ in all UCB-based algorithms.
\subsection{Compute}
 The experiments are conducted on AMD Ryzen 9 5900HX CPU @ 3.3GHz, and 16 GB of RAM and we use single thread.
 \subsection{Implementation}
 The experiments were implemented in Python. The major libraries used were GPyTorch and BoTorch. All experiments use a squared exponential kernel with ARD lengthscales.

\end{document}